\documentclass[final,3p,times]{elsarticle}
\usepackage{graphicx}
\usepackage{epsfig}
\usepackage{amssymb}
\usepackage{mathrsfs}
\usepackage{bm}
\usepackage{color}
\usepackage{amsmath}
\usepackage{mathrsfs}
\usepackage{geometry}
\usepackage{graphicx}
\usepackage{subfigure}
\usepackage{float}
\usepackage{booktabs}
\usepackage{multirow}
\usepackage{epstopdf}
\usepackage[ruled]{algorithm2e}
\biboptions{numbers,sort&compress}  
\usepackage[colorlinks,
            linkcolor=blue,
            anchorcolor=blue,
            citecolor=blue]{hyperref}

\usepackage{threeparttable}

\journal{...}
\usepackage{CJK}
\begin{document}
\begin{CJK*}{GB}{gbsn}
\begin{frontmatter}

\title{Fairness and Accuracy in Federated Learning}
\author[adr1,adr2,adr3]{Wei Huang}
\ead{huangweifujian@my.swjtu.edu.cn}
\author[adr1,adr2,adr3]{Tianrui Li\corref{cor1}}
\ead{trli@swjtu.edu.cn}
\cortext[cor1]{Corresponding author}
\author[adr1,adr2,adr3]{Dexian Wang}
\ead{xiaoke92@foxmail.com}
\author[adr1,adr2,adr3]{Shengdong Du}
\ead{sddu@swjtu.edu.cn}
\author[adr4,adr5,adr2]{Junbo Zhang}
\ead{msjunbozhang@outlook.com}

\address[adr1]{School of Information Science and Technology, Southwest Jiaotong University, Chengdu 611756, China}
\address[adr2]{Institute of Artificial Intelligence, Southwest Jiaotong University, Chengdu, China}
\address[adr3]{National Engineering Laboratory of Integrated Transportation Big Data Application Technology, Southwest Jiaotong University, Chengdu, China}
\address[adr4]{JD Intelligent Cities Business Unit, JD Digits, Beijing, China}
\address[adr5]{JD Intelligent Cities Research, Beijing, China}

\begin{abstract}
  In the federated learning setting, multiple clients jointly train a model under the coordination of the central server, while the training data is kept on the client to ensure privacy. Normally, inconsistent distribution of data across different devices in a federated network and limited communication bandwidth between end devices impose both statistical heterogeneity and expensive communication as major challenges for federated learning. This paper proposes an algorithm to achieve more fairness and accuracy in federated learning (FedFa). It introduces an optimization scheme that employs a double momentum gradient, thereby accelerating the convergence rate of the model. An appropriate weight selection algorithm that combines the information quantity of training accuracy and training frequency to measure the weights  is proposed. This procedure assists in addressing the issue of unfairness in federated learning due to preferences for certain clients. Our results show that the proposed FedFa algorithm outperforms the baseline algorithm in terms of accuracy and fairness.
\end{abstract}
\begin{keyword}
federated learning; weighting strategies; momentum gradient; information quantity
\end{keyword}
\end{frontmatter}

\section{Introduction}
The integration of the internet of things (IoT), edge computing, and artificial intelligence (AI) requires the safe and legal management of distributed big data. For example, many smartphones store private data, which needs to be integrated for training. Alternatively, in the case of automated vehicles, it is essential for each vehicle to be able to process large amounts of information locally using machine learning models, while working in concert with other vehicles and computing centers \cite{yang2019federated}. In recent years, despite their success in overcoming technical and computational limitations, these technologies still face problems closely related to data security. Generally, service providers collect the data from users to a central server to train machine learning models, including deep neural networks. The centralized machine learning approach leads to serious practical problems, such as high communication costs, large consumption of device batteries, and the risk of violating the privacy and security of user data. At the same time, there is a worldwide trend toward tighter management of user data privacy and security, which is reflected in the recent introduction of the General Data Protection Regulation (GDPR) in the European Union \cite{kotsios2019analysis}.

The establishment of data security regulations will clearly contribute to a more civilized society, but will also pose new challenges to the data handling procedures commonly used in AI. In this new legislative environment, it is becoming increasingly difficult to collect and share data between different organizations. In addition, the sensitive nature of certain data (e.g., financial transactions and medical records) prohibits the collection, fusion, and use of data, and forces data to exist in isolated data silos maintained by data owners \cite{2019Federated}. As a result, a way has been sought to train machine learning models without having to centralize all data into a central storage point.

The concept of federated learning was first introduced by McMahan et al. \cite{mcmahan2017communication} in 2016. Federated learning has received a lot of attention for its ability to train large-scale models in a decentralized manner without direct access to user data. It helps protect users' private data from centralized collection. In contrast to distributed machine learning, federated learning aims to process heterogeneous and unbalanced data from a variety of real-world applications (e.g., apps in smartphones).

Federated Learning is a machine learning framework that effectively helps multiple organizations perform modeling of data while meeting user privacy, data security, and government regulatory requirements. Federated learning, as a distributed machine learning paradigm, can effectively solve the problem of data silos, allowing participants to jointly model without sharing data, which can technically break down data silos and enable AI collaboration. Federated networks consist of a large number of devices, so communication speeds in the network are many orders of magnitude slower than local computing speeds. Devices often generate and collect data in a way that is not identically distributed across the network, and the amount of data across devices can vary greatly. This data generation violates the assumption of independent identical distributions commonly used in distributed optimization and increases the complexity of modeling, analysis, and evaluation.

Compared with other machine learning paradiagms, federated learning is subject to the following challenges \cite{mcmahan2017communication}:

1)\textbf{\emph{Non-Independent Identical Distribution (Non-IID):}} The distribution of data in each device is not representative of the global data distribution, i.e., the categories in each device are incomplete. For example, there are 100 categories of images in the full set, and in one device are landscape images, while in another device are people or plant images. The former is one distribution and the latter is another. That is the Non-IID in federated learning. Conversely, if there are 100 categories of pictures in a certain device, and the other devices also have these 100 categories of pictures, then they are identically distributed. In real daily life, data samples distributed across different platforms are not independently and homogeneously distributed. When dealing with Non-IID data, the training complexity of the federated learning model may be greatly increased.

2)\textbf{\emph{Unbalanced:}} The data amount on the clients may be different, because some clients produce a lot of data for model learning and some clients produce less. For example, a device has one hundred pictures in the landscape category and ten pictures in the animal category, while a device has ten thousand pictures in the landscape category and one thousand pictures in the animal category. The difference in the number of samples per device or in the number of samples per category is generally described as unbalanced. Unbalanced data also creates the problem of heterogeneity, which leads to an increase in the training complexity of the federated learning model.

3)\textbf{\emph{Massively distributed:}} The size of the clients participating in federated learning is much larger than the average number of examples per client. The clients for distributed learning are the compute nodes in a single cluster or data center, which are typically 1$\sim$1000 clients. Cross-silo federated learning is commonly 2$\sim$100 clients. While cross-device federated learning uses massive parallelism and can reach $10^{10}$ clients.

4)\textbf{\emph{Limited communication:}} Clients that participate in model learning are frequently offline or on slow or expensive connections. Participating in federated learning may be mobile devices such as cell phones and tablets, which may be disconnected from the central server at any time due to the habits of users and the unstable network environment where they are located. Moreover, the devices in federated learning may be distributed in different geographical locations, and they are generally connected to the central server remotely, and the communication cost is higher due to the network bandwidth of different devices. Therefore, with a large number of clients participating in federated learning, communication bandwidth is increasingly becoming a bottleneck.

In this work, we propose FedFa, a federated optimization algorithm with a double momentum gradient method, and an appropriate weighting strategy based on the information of the frequency of participation or learning accuracy. We introduce momentum gradient descent in both client and server to help the gradient converge faster, and also weight clients appropriately using the information quantity about the accuracy and frequency of participation in the training. As a result, our proposed method has faster convergence and stability in comparison to traditional federated learning.

The remainder of the paper is organized as follows. In \textcolor[rgb]{0.00,0.00,1.00}{Section 2}, we provide a related work on federated learning. The detail of the proposed algorithm, FedFa, especially the double momentum gradient method, and appropriate weighting strategies are described in \textcolor[rgb]{0.00,0.00,1.00}{Section 3}. In \textcolor[rgb]{0.00,0.00,1.00}{Section 4}, we provide a thorough empirical evaluation of FedFa on a suite of synthetic and real-world federated datasets. Finally, conclusions are drawn in \textcolor[rgb]{0.00,0.00,1.00}{Section 5}.

\section{Related Work}
Federated learning faces some challenges includes the joint optimization of heterogeneous data and expensive communication. The Federal Average Method (FedAvg), as the fundamental framework under the FL setup, has been improved by numerous researchers \cite{Secure,li2020developing,averaging}.

\textbf{\emph{Heterogeneity}} is a fundamental problem for federated learning. If heterogeneity is not addressed, the accuracy of the model cannot be improved, making it impossible to put federated learning into practical use. Heterogeneity in federated learning encompasses both systematic and statistical heterogeneity \cite{mohri2019agnostic,huang2018loadaboost,corinzia2019variational,nishio2019client}. Zhao et al. \cite{zhao2018federated} argued that the loss of accuracy of federated learning on Non-IID data can be explained by weight divergence, which improves the training of non-iid data by introducing EMD (earth move distance) distances and sharing a small portion of global data between clients. While this approach does allow for the creation of more accurate models, it has several drawbacks, the most critical of which is that we typically cannot assume the availability of such a public dataset. Sattler et al. \cite{Clustered} proposed the Cluster Federal Learning (CFL) algorithm, which allows clients with similar data to cooperate with each other while minimizing interference between clients with different data. CFL is flexible enough to handle client populations that change over time and can be implemented in a privacy preserving manner. However, the time complexity of training the CFL algorithm during splitting is high and the communication burden increases. Huang et al. \cite{huang2018loadaboost} borrowed the idea from the AdaBoost algorithm in \cite{1995A}. According to the median value of loss for each client in the previous round, the number of epochs trained on the local model for the client in the current iteration round is dynamically adjusted to solve the problem of inconsistent data distribution. Li et al. \cite{li2018federated} proposed a broader framework based on FedAvg, FedProx, which is capable of handling heterogeneous federated data, while maintaining similar privacy and computational advantages. The framework does not depend on the particular solver used on each device and can improve the robustness and stability of convergence in heterogeneous federated networks. However, the proximal term constraint is isotropic and is not personalized for solving client heterogeneity problems in federated learning. The team also presented \cite{li2019fair} the q-FFL method, which constructs a novel goal that can improve the fairness of the accuracy distribution in federated learning. Simply minimizing the total loss in the network may benefit or disadvantage some devices, so q-FFL can achieve a more uniform accuracy distribution across devices, significantly reducing variance while ensuring average accuracy.

\textbf{\emph{Communication}} optimization problems for federated learning can be categorized mainly into communication bandwidth-based, communication latency-based, and network topology-based \cite{yao2019federated,chen2016revisiting,zhang2015deep,chen2019communication}. The local update method can reduce the total number of rounds of communication, Yao et al. developed the FedMMD \cite{yao2018two} and FedFusion \cite{yao2019towards} algorithms, where FedMMD incorporates MMD loss in the objective function of the local model and FedFusion fuses the global model in the training of the local model on the client side. Experiments have shown that both algorithms converge faster, with higher accuracy and fewer communications than the FedAvg algorithm. Model compression schemes (e.g., sparsification, subsampling, and quantization) can significantly reduce the message size per round of communication. Some work has provided practical strategies in federated settings, such as forcing the update model to become sparse and low-ranked, performing quantization using structured random rotation \cite{konevcny2016Strategies}, using lossy compression and random inactivation to reduce server-to-device communication \cite{caldas2018expanding}, and applying Golomb lossless coding \cite{sattler2019robust}. Ribero et al. \cite{ribero2020communication} derived an optimal sampling strategy to reduce communication by modeling the variation in customer weights through the Ornstein-Uhlenbeck process in a communication-constrained environment. Li et al. \cite{li2020blockchain} proposed a decentralized federated learning framework based on blockchain, i.e., a Blockchain-based Federated Learning framework with Committee consensus (BFLC). Decentralized algorithms in federated learning could theoretically reduce the high communication costs on a central server.

There are also many other researches on federated learning, e.g., federated multi-task learning, federated meta-learning, personalized Federated Learning and federated multi-center learning, etc. Shoham et al. \cite{shoham2019overcoming} proposed the FedCur framework, which combines the idea of multi-task learning to address how to learn a task in federated learning without interfering with different tasks learned on the same model. Smith et al. \cite{smith2017federated} introduced the MOCHA framework that connects each task with different parameters and learns the independence of each device by adding loss terms, while modeling task correlations through multi-task learning using shared representations. MOCHA uses a primal pairwise formulation to solve optimization problems, which is limited to convex objectives, and has limited ability to extend large-scale networks and is not suitable for deep networks. Jiang et al. \cite{jiang2019improving} argued that the FedAvg method proposed by McMahan \cite{mcmahan2017communication} has many similarities with MAML \cite{finn2017model} and can be interpreted in terms of meta-learning algorithms. Dinh et al. \cite{t2020personalized} put forward a personalized federated learning scheme, which introduces moreau envelopes optimization based on client-side loss functions.  With this scheme, the client can not only construct a global model as in classical federated learning, but also use the global model to optimize its personalized model. Xie et al. \cite{xie2020multi} developed a novel multi-center aggregation method to address the Non-IID challenges of federated learning , and presented the federated stochastic expectation maximization (FeSEM) to solve the optimization problem of the proposed objective function.  However, the influence of prior values of multiple centers is significant, and the initialized values are uncertain for the optimization process.

The heterogeneous nature of federated learning leads to the possibility of preferences for certain clients in the training process, i.e., there may be unfairness. We leverage the accuracy and frequency of our clients' involvement in the training process to select the appropriate weights for each round of server aggregation. One of the best ways to reduce network overload is to quickly and stably converge the target accuracy of the learning model in federated learning. Therefore, we introduce the momentum gradient descent method for the client and server, respectively.

\section{Fairness and accuracy in federated learning (FedFa) }
\begin{figure}[htb]
  \centering
  \includegraphics[width=8cm]{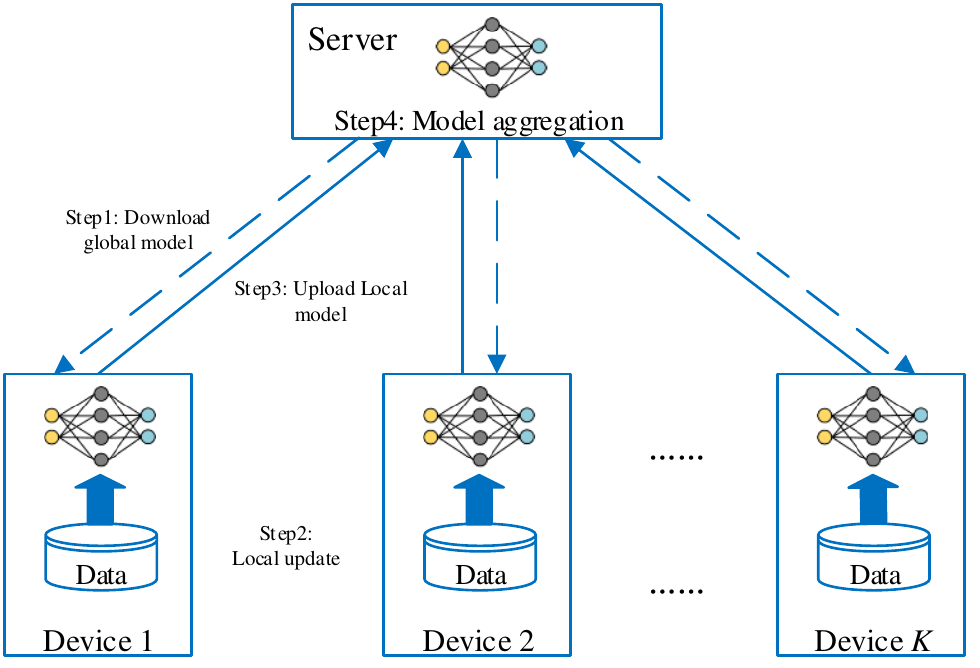}\\
  \caption{Framework of a federated learning.}\label{fig: framework}
\end{figure}

In this section, we will introduce fairness and accuracy in federated learning (FedFa). It is presented in two parts: the double momentum gradient approach and the appropriate weighting strategies. We begin with an introduction to the basic concepts of federated learning. Figure \ref{fig: framework} illustrates a framework for federated learning.

In FedAvg, the framework is composed of a central server $PS$ and $K$ clients. As shown in Figure \ref{fig: framework}, the process of the entire algorithm has the following steps for a $T$-round training loop: first, the server randomly selects a subset $K \ll N$ of the total devices. Then, the initialized parameter $w_{t}$ is sent to these devices, and each device updates the parameter using the local dataset and sends it back to the server. Finally, the server aggregates and averages the parameters sent by these devices.

The pseudo code of FedAvg is provided in Algorithm \ref{FedAvg}.

\begin{algorithm}[H]
\caption{Federated Averaging \cite{mcmahan2017communication} (FedAvg)}\label{FedAvg}
\KwIn{input parameters $K$, $T$, $\eta$, $E$, $w^{0}$, $N$, $p_{k}$, $k=1, ..., N$}
\textbf{for} $t=0, ..., T-1$ \textbf{do}\; 
\quad \ Server selects a subset $S_{t}$ of $K$ devices at random (each device $k$ is chosen with probability $p_{k}$)\;
\quad \ Server sends $w^{t}$ to all chosen devices\;
\quad \ Each device $k\in S_{t}$ updates $w^{t}$ for $E$ epochs of SGD on $F_{k}$ with step-size $\eta$ to obtain $w_{k}^{t+1}$\;
\quad \ Each device $k\in S_{t}$ sends $w_{k}^{t+1}$ back to the server \;
\quad \ Server aggregates the $w$'s as $w^{t+1}=\Sigma_{k\in S_{t}} \frac{n_{k}}{n} \ w_{k}^{t+1}$ \;
\textbf{end for}
\end{algorithm}

\subsection{Double momentum gradient method}
The ordinary gradient descent method is sufficient to solve conventional problems, such as linear regression. However, when the problem becomes complex, the ordinary gradient descent method faces many limitations. Specifically, for the ordinary gradient descent formula $w=w-\eta \Delta w$, where $\eta$ denotes the learning rate and is the step size of the gradient adjustment on each time step. The gradient will be smaller when close to the optimal value. Since the learning rate is fixed, the ordinary gradient descent method will converge slower and sometimes even fall into a local optimum. If the historical gradient is taken into account, it will lead the parameters to converge faster towards the optimal value. The basic idea behind the momentum gradient descent method is to calculate the exponentially weighted average of the gradient and use that gradient to update the weights. If the current gradient descends in the same direction as the last update, the last update can act as a positive acceleration to the current search. Conversely, the last update can act as a deceleration to the current search. Thus, the formula for updating the client's parameters is as follows:

\label{equ:momentum1}
\begin{align}
&m = \gamma m+\eta \Delta w_{k}^{t+1},\\
&w_{k}^{t+1} = w_{k}^{t}-m,
\end{align}
where $m$ is the momentum term. $\gamma$ denotes the influence of the historical gradient, and the larger $\gamma$ is, the greater the influence of the historical gradient on the present. $t$ represents the round of updates.

It is normal for it to use momentum gradient descent in order for the client's model to accelerate convergence. When the server side aggregates the client's gradients, we believe that the historical gradients need to be considered as well. Based on the idea of momentum gradient, it will lead to faster convergence of the server aggregated parameters towards the optimal value and reduce communication. However, there is ordinarily no data on the server side in horizontal federated learning. Therefore, this paper takes the following formula of the server's parameter updates during model training:
\label{equ:momentum2}
\begin{align}
&w^{t+1} = \frac{1}{k}\Sigma_{k\in S_{t}} \ w_{k}^{t+1},\\
&\Delta w = w^{t+1}-w^{t},\\
&m = \gamma m+(1-\gamma) \Delta w,\\
&w^{t+1} = w^{t+1}-\eta m, \label{equ:momentum2.4}
\end{align}
where $\Delta w$ is the server-side global gradient. We approximate this gradient on the server side by computing the difference between the latest converged global model ($t+1$ round) and the previous global model ($t$ round). Generally, in federated learning, the data between clients is Non-IID and the selection of clients to participate in the training is random each round. Therefore, we can wait for $b$ rounds before using an approximate method of updating the momentum gradient on the server side ($b$ can take 3, 5, etc.).  The purpose of this is to allow the server-side to better capture the impact of the historical gradient. Then, the formula (\ref{equ:momentum2.4}) will be changed in the following form:
\label{equ:momentum3}
\begin{align}
&w^{t+1} = w^{t+1}-\eta m \ \ (t+1=bn,n\in N^{*}). \label{equ:momentum3.1}
\end{align}
The value of $t+1$ is an integer multiple of $b$. It means that the serve uses the formula (\ref{equ:momentum3.1}) to update every $b$ round. A proper value of $b$ will make the global model consider the historical gradient more comprehensively, reduce oscillations, and ensure efficiency and correct convergence.

\subsection{Appropriate weighting strategies}
Generally, the number of devices in a federated network is large, which ranges from hundreds to millions. While the general goal of federated learning is to fit the model through some kind of empirical risk minimization objective to the data that is generated by the network of devices. Simply minimizing the average loss in such a large network may overly favor model performance on some devices. Although the average accuracy may be high, the accuracy of each device in the network may not always be very well. The situation is made worse by the reality that data is usually heterogeneous between devices in terms of size and distribution. Therefore, in this work, we design a weighting strategy to encourage a more fair distribution of model performance across devices in the federated network.

In FedAvg, it commonly implements a weighted aggregation strategy for each local model based on the size of the training data on the client. It is defined by
\label{equ:weight1}
\begin{align}
w^{t+1}=\Sigma_{k\in S_{t}} \frac{n_{k}}{n} \ w_{k}^{t+1}, \label{equ:weight11}
\end{align}
where $n_{k}$ is the sample size of $k$-th client and $n$ is the total number of training samples. Obviously, the larger the $n_{k}$, the greater the contribution of the local model on client $k$ to the central model. Temporally weighted aggregation algorithm was proposed by \cite{chen2019communication} to make federated learning more efficient in communication. It achieves faster convergence of learning accuracy compared to traditional federated learning. But it only takes into account the latest round of learning for each user.

Aiming to improve learning speed and stability, we propose an appropriate weighting strategy in federated learning. When dealing with Non-IID data, the quality of the client's data may be different and the corresponding training accuracy may be also varied. In this case, the gradient of the local model should be processed differently in the aggregation step. And since the selection of clients to participate in the training is random in each round, the number of times each client participates in the training is also diverse. Therefore, we use the training accuracy $Acc_{i}$ and the number of training participation $f_{i}$ as weighting factors. At each training round, the client trains the model locally and sends the training accuracy and the number of training participation along with the gradient to the server. The server then performs a weighted aggregation based on the training accuracy and number of training participation of the clients. Since the amount of data for each client is constant, but the adaptability of the model to the client's data varies from round to round, and the number of times the client participates in the training keeps changing. Therefore, we use the factors of change as a measure of weighted strategy in hopes of creating a more fair and accurate model in federated learning.

After the client send information to the server in each round, the server normalizes the client's $Acc_{i}$ and $f_{i}$ as follows:
\label{equ:weight2}
\begin{align}
&Acc_{i}=\frac{Acc_{i}}{\sum_{i=1}^{K} Acc_{i}},\\
&f_{i}=\frac{f_{i}}{\sum_{i=1}^{K} f_{i}}.
\end{align}

A weighting approach that simply utilizes only the client's data size does not ensure that federated learning is fair in the training process. The amount of information available to the client will vary as a result of the training accuracy and the frequency of participation.

We utilize the information quantity to measure weight. Due to the complexity of the local dataset, there can be significant differences in training accuracy between clients. In general, the lower the training accuracy, the more information the client has to learn. This is also intended to address the preference for certain clients in the federated learning process and to achieve a more fair result.
\label{equ:weight3}
\begin{equation}
\label{equ:weight31}
Acc_{i}\_inf=\left\{
\begin{aligned}
&-log_{2} Acc_{i}, & Acc_{i}\neq 0, \\
&-log_{2} Acc_{i}+c, & Acc_{i}= 0,
\end{aligned}
\right.
\end{equation}
where $c$ is a very small constant close to 0. It is to avoid the antilogarithm of logarithms being 0.

The more rounds a client participates in training, the more information it has. On the other hand, the less information it has.
\label{equ:weight4}
\begin{equation}
f_{i}\_inf=\left\{
\begin{aligned}
&-log_{2} (1-f_{i}), & 1-f_{i}\neq 0, \\
&-log_{2} (1-f_{i}+c), & 1-f_{i}= 0,
\end{aligned}
\right.
\end{equation}
where $c$ has the same function as in formula \ref{equ:weight31}. Then, the information quantity for $Acc_{i}\_entropy$ and $f_{i}\_entropy$ can be normalized as follows:
\label{equ:weight5}
\begin{align}
&Acc_{i}\_inf=\frac{Acc_{i}\_inf}{\sum_{i=1}^{K} Acc_{i}\_inf},\\
&f_{i}\_inf=\frac{f_{i}\_inf}{\sum_{i=1}^{K} f_{i}\_inf}.
\end{align}

We apply weights based on the frequency and training accuracy of learning to participate in the model aggregation process.
\label{equ:weight6}
\begin{align}
&weight_{i}=\alpha Acc_{i}\_inf+\beta f_{i}\_inf,
\end{align}
where, $\alpha$+$\beta$=1. The formula for model aggregation is proposed based on the client's model accuracy and frequency of participation in learning. $\alpha$ and $\beta$ are the proportional effects of each method on the aggregated model. The pseudo code of FedFa is provided in Algorithm \ref{FedFa}.

\begin{algorithm}[H]
\caption{ Fairness and Accuracy Federated Learning (FedFa)}\label{FedFa}
\KwIn{input parameters $K$, $T$, $\eta$, $E$, $w^{0}$, $N$, $p_{k}$, $k=1, ..., N$}
\textbf{for} $t=0, ..., T-1$ \textbf{do} :\\ 
\quad \ Server selects a subset $S_{t}$ of $K$ devices at random (each device $k$ is chosen with probability $p_{k}$)\;
\quad \ Server sends $w^{t}$ to all chosen devices\;
\quad \ Client :\\
\quad \ \quad \ \quad \ Each device $k\in S_{t}$ updates $w^{t}$ for $E$ epochs of gradient descent with Momentum on $F_{k}$ with step-size $\eta$ to obtain $w_{k}^{t+1}$\;
\quad \ \quad \ \quad \ Each device $k\in S_{t}$ sends $w_{k}^{t+1}$, $Acc_{k}^{t+1}$, $f_{k}^{t+1}$ back to the server \;
\quad \ Server :\\
\quad \ \quad \ \quad \ calculate the amount of information for $Acc_{k}\_inf$ and $f_{k}\_inf$ \;
\quad \ \quad \ \quad \ update the $weight$ as: $weight_{k}=\alpha Acc_{k}\_inf+\beta f_{k}\_inf$ \;
\quad \ \quad \ \quad \ aggregate the $w$'s as $w^{t+1}=\Sigma_{k\in S_{t}} weight_{k} \times \ w_{k}^{t+1}$ \;
\quad \ \quad \ \quad \ calculate the difference between two rounds of $w$ as a gradient: $\Delta w = w^{t+1}-w^{t}$ \;
\quad \ \quad \ \quad \ calculate the momentum term: $ m = \gamma m+(1-\gamma) \Delta w$ \;
\quad \ \quad \ \quad \ update the $w$ as: $w^{t+1} = w^{t+1}-\eta m$ \;
\textbf{end for}
\end{algorithm}
\section{Evaluation}
In this section, we present the empirical results of the proposed FedFa algorithm. In Section 4.1, we describe the experimental setup including the datasets used. Then, we demonstrate the effectiveness of FedFa for heterogeneous networks in Section 4.2. And the improved fairness of FedFa is illustrated in Section 4.3. Finally, in Section 4.4, we discuss the effect of parameters. All code, data and experiments will be available by sending the email after the paper is accepted.

\subsection{Experimental Design}
\textbf{\emph{Federated Datasets}}. Our model is compared to baselines on one synthetic and four real datasets  to validate our effectiveness and fairness. These datasets are drawn from previous federated learning work \cite{mcmahan2017communication} \cite{li2018federated}.  Table \ref{dataset} summarizes the statistical information for the four real datasets. Details of all datasets are also presented below.

\begin{table*}[!ht]
\centering
\caption{Statistics of four real federated datasets.}\label{dataset}
\resizebox{0.45\textwidth}{!}{
\begin{tabular}{lllll}
  \toprule
  Dataset&Devices&Samples & \multicolumn{2}{c}{Samples/device}\\
  \cmidrule{4-5}
   & & &mean&stdev\\ \midrule
  MNIST&1,000&69,035&69&106\\
  FEMNIST&200&18,345&92&159\\
  Sent140&772&40,783&53&32\\
  Shakespeare&143&517,106&3,616&6,808\\
  \bottomrule
\end{tabular}}
\end{table*}

$\bullet$ \emph{Synthetic: }These are three Non-IID datasets which set $(\alpha,\beta)=(0,0), (0.5,0.5)$ and $(1,1)$, respectively. And one IID data which set the same $W,b\thicksim \mathcal{N}(0,1)$ on every device. In addition, $x_{k}$ follows the same distribution $\mathcal{N}(v,\sum)$, where each element in the mean vector is zero and $\sum$ is diagonal to $\sum _{j,j}=j^{-1.2}$. There are 30 devices in total for all synthetic datasets, and the number of samples on each device follows the power law. The increase in $\alpha$ and $\beta$ indicates that the degree of heterogeneity of the dataset increases.

$\bullet$ \emph{MNIST: }The MNIST \cite{Gradient} dataset is an image classification of handwritten digits 0-9. The dataset input to the model of multinomial logistic regression is a 784-dimensional $(28 \times 28)$ flattened image and the output is a class label between 0 and 9. Previous studies have distributed the data across 1000 devices. In order to simulate statistical heterogeneity, each device has only 2 digits per sample and the number of samples follows the power law.

$\bullet$ \emph{FEMNIST: }FEMNIST is a federated dataset based on EMNIST with a total of 200 devices. The EMNIST dataset \cite{EMNIST} is an image classification problem with 62 classes. In order to form a heterogeneous distribution of devices, the dataset takes 10 lowercase letters (`a'-`j') as subsamples from EMNIST and assigns only 5 classes to each device. A flattened 784-dimensional $(28 \times 28)$ image was used as input to the model of multinomial logistic regression, and the output was a class label between 0 and 9.

$\bullet$ \emph{Sent140: }This dataset is a collection of Sentiment 140 \cite{go2009twitter} (Sent140) tweets for the text sentiment analysis task. Each Twitter account corresponds to one device. We model it as a binary classification problem. The model takes a sequence of 25 characters as input, embeds each word in a 300-dimensional space using pre-trained Glove \cite{2014Glove}, and outputs a binary tag after two LSTM layers and a dense join layer.

$\bullet$ \emph{Shakespeare: }This is a dataset built from \emph{The Complete Works of William Shakespeare} \cite{mcmahan2017communication}. The model treats a sequence of 80 characters as input, embeds each character in the 8-dimensional space of learning, and outputs a character after two LSTM layers and a tightly connected layer. Each speaking character in a play represents a different device.

\textbf{\emph{Implementation.}} We implement all the code in TensorFlow \cite{abadi2016tensorflow} Version 1.10.0 to simulate a federated network with one server and $K$ devices. We randomly divide the data on each local device into 80\% of the training set and 20\% of the test set. For all experiments on all datasets, we fix the number of selected devices per round to 10.  For all synthetic data experiments in FedFa, the learning rate is 0.0001. The server and client momentum factors are (0.5, 0.9) or (0.5, 0.5), respectively. For a fair comparison with all synthetic data experiments in FedAvg and FedProx, the learning rate is 0.01, 0.001 and 0.0001. For MNIST, FEMNIST, Sent140 and Shakespeare in FedFa, the learning rate is 0.0003, 0.0003, 0.0001, 0.08. For MNIST, FEMNIST, Sent140 and Shakespeare in FedAvg and FedProx, we use the learning rates of 0.03, 0.003, 0.3, and 0.8.  

\subsection{Effects of double momentum gradient}

\begin{figure}[htbp]
\centering
\subfigure[Testing Accuracy]{
\includegraphics[width=14cm]{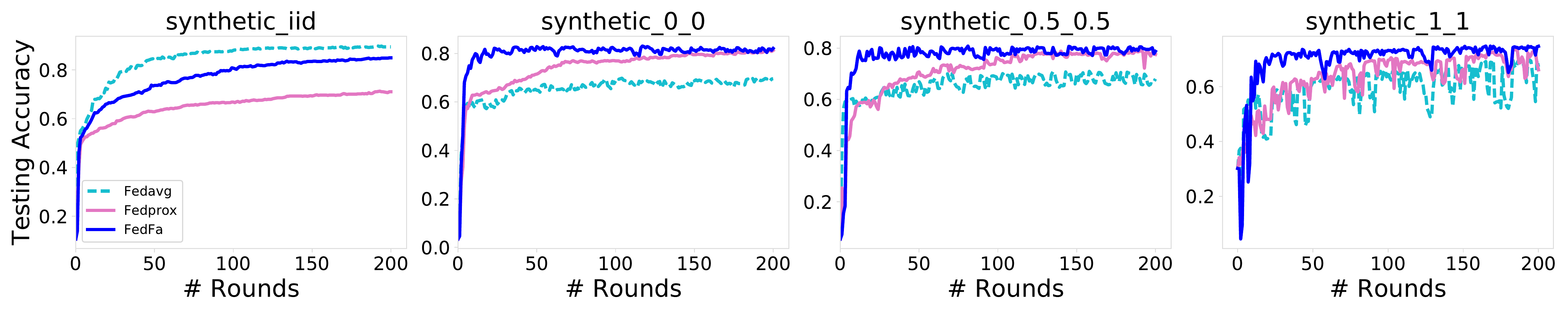}
\label{fig: accuracy-01-0001}
}

\quad
\subfigure[Training Loss]{
\includegraphics[width=14cm]{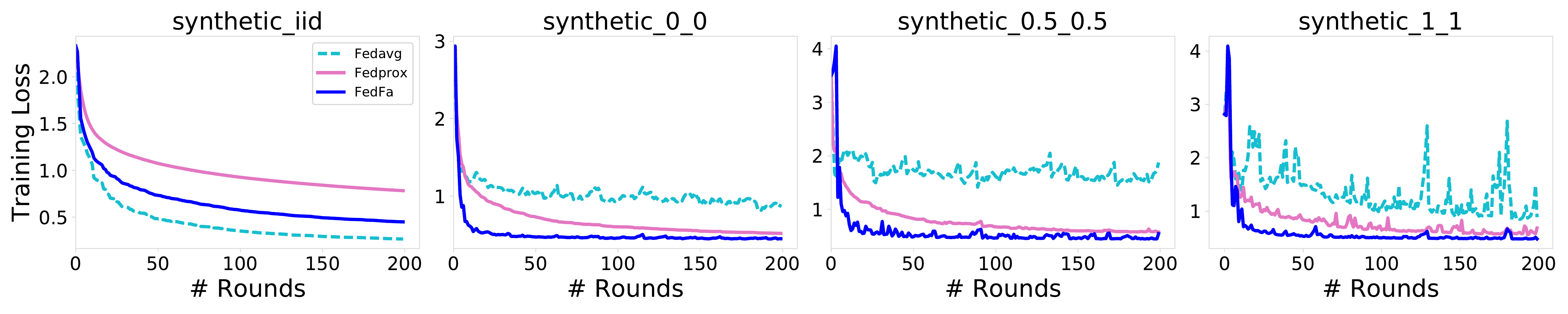}
\label{fig: loss-01-0001}
}
\caption{The Training Loss and Test Accuracy of Experiments on Synthetic Datasets with Four Different Data Distributions (From left to right, the data are increasingly heterogeneous). The learning rate for FedAvg and FedProx is 0.01. For the two synthetic data sets on the left, the momentum factors for the client and server side of FedFa are 0.9 and 0.5, respectively. For the two synthetic data sets on the right, the momentum factors for the client and server side of FedFa are 0.5 and 0.5, respectively.}
\label{fig: accuracyloss-01-0001}
\end{figure}

We use four synthetic datasets to study how statistical heterogeneity affects convergence (fixing $E$ to be 20), and to demonstrate the effectiveness of our double dynamic mechanism. In Figure \ref{fig: accuracyloss-01-0001}, we compare the FedFa algorithm (which only has the momentum mechanism without applying the weighting strategy) with FedAvg and FedProx with a learning rate of 0.01 (where the parameter settings of FedProx and FedAvg are strictly referenced to the settings in \cite{li2018federated}).

Both the test accuracy and training loss in Figure \ref{fig: accuracyloss-01-0001} show that the convergence effect of FedAvg becomes gradually unstable and fluctuates as the data heterogeneity increases. However, FedProx can cope with the problem of statistical heterogeneity. It can be seen from the experiments that FedProx converges better than FedAvg as the data becomes more heterogeneous. Obviously, FedAvg performs well in the case of IID data. However, in practice, it is difficult to have perfect IID data. By applying our proposed double-momentum mechanism, FedFa can not only improve the convergence speed in the case of IID data, but also speed up the convergence speed for heterogeneous data. Since FedFa takes into account the historical gradient information at both server and client sides, this helps the algorithm to converge better.

\begin{figure}[htbp]
\centering
\subfigure[Testing Accuracy]{
\includegraphics[width=14cm]{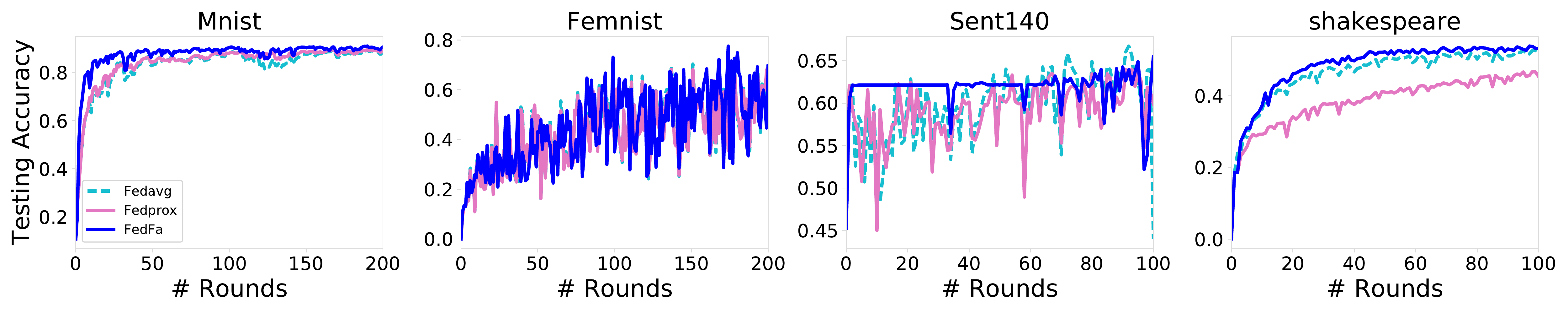}
\label{fig: accuracy-real}
}

\quad
\subfigure[Training Loss]{
\includegraphics[width=14cm]{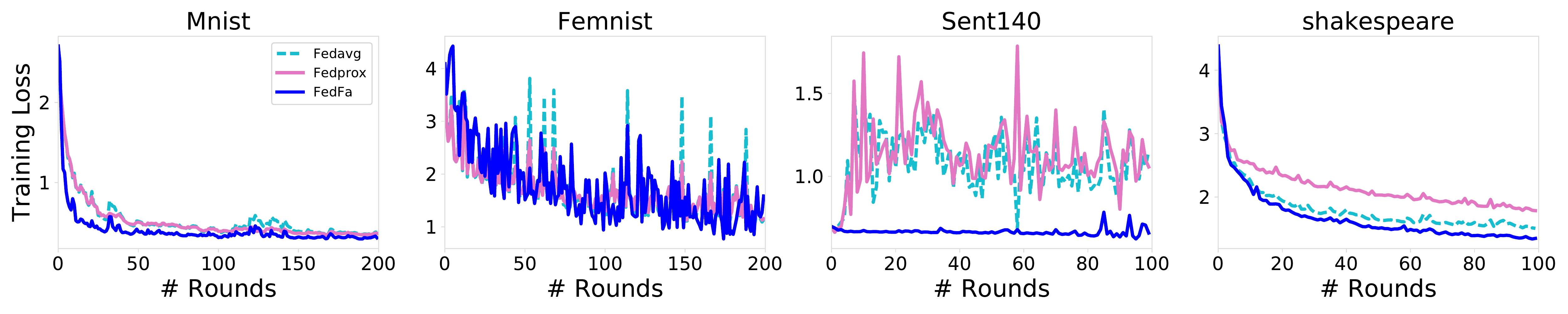}
\label{fig: loss-real}
}
\caption{The Training Loss and Test Accuracy of Experiments on four real Datasets. For the three datasets (Mnist, Femnist, Shakespeare), the momentum factors for the client and server side of FedFa are 0.5 and 0.5, respectively. For the dataset of Sent140, the momentum factors for the client and server side of FedFa are 0.1 and 0.5}
\label{fig: real}
\end{figure}

We also study the effects of different momentum factors on the client and server side. The synthetic dataset with the same distribution (synthetic\_iid) and the less heterogeneous synthetic dataset (synthetic\_0\_0) use the momentum factors of the client and the server of 0.9 and 0.5. Since for the less heterogeneous dataset, speeding up the convergence of the client does not lead to global gradient deviations. For the more heterogeneous synthetic dataset (synthetic\_0.5\_0.5 and synthetic\_1\_1),  we set the momentum factors for the client and server sides to 0.5 and 0.5. We argue that this will reduce the oscillations on the server side when aggregating the gradients during the heterogeneous data training process.

As shown in Figure \ref{fig: real}, in order to validate the double momentum mechanism, we also conduct experiments on four real datasets. The figure shows the test accuracy and training loss of FedAvg, FedProx, and FedFa on these real datasets. It can be seen that the double momentum mechanism can also be worked in real datasets.

\begin{figure}[htbp]
\centering
\subfigure[Testing Accuracy]{
\includegraphics[width=13.7cm]{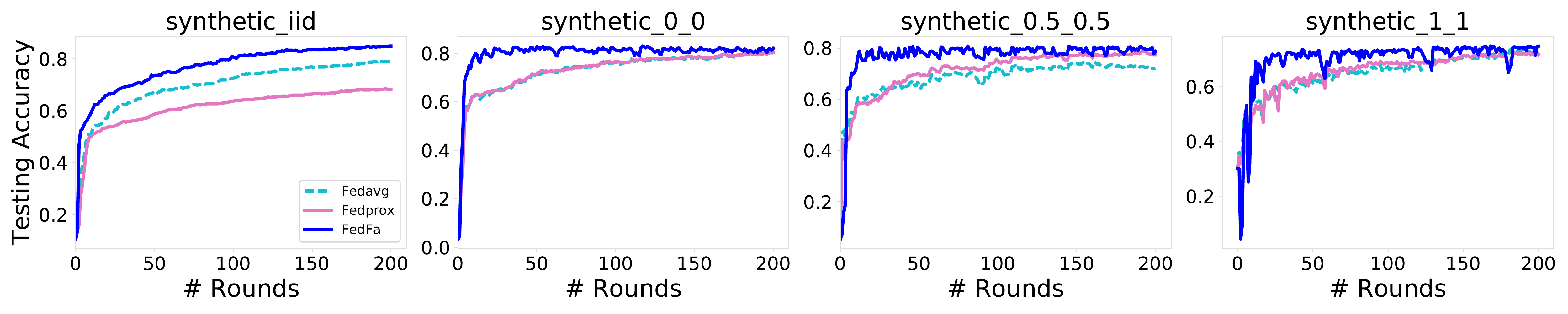}
\label{fig: accuracy-001-0001}
}

\quad
\subfigure[Training Loss]{
\includegraphics[width=13.7cm]{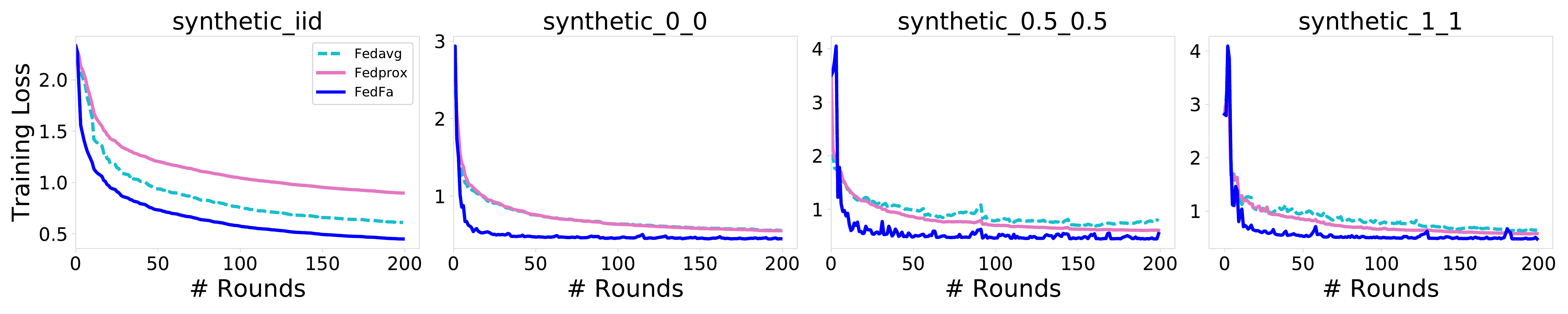}
\label{fig: loss-001-0001}
}
\caption{The Training Loss and Test Accuracy of Experiments on Synthetic Datasets with Four Different Data Distributions (From left to right, the data are increasingly heterogeneous). The learning rate for FedAvg and FedProx is 0.001. All settings of FedFa are kept fixed.}
\label{fig: accuracyloss-001-0001}
\end{figure}

For a fair comparison, we explore FedAvg and FedProx on synthetic datasets at learning rates of 0.001 and 0.0001. As shown in Figure \ref{fig: accuracyloss-001-0001}, the learning rate of FedAvg and FedProx is 0.001. We can see that despite reducing the learning rate of FedAvg and Fedprox, the experimental test accuracy and loss can reduce oscillations during training, but will impair convergence. In addition, as can be seen in Figure \ref{fig: accuracyloss-001-0001}, our method still works better than their results. Furthermore, our method performs better than FedAvg and FedProx on the synthetic dataset of IID when these two algorithms reduce the learning rate.

\begin{figure}[htbp]
\centering
\subfigure[Testing Accuracy]{
\includegraphics[width=13.7cm]{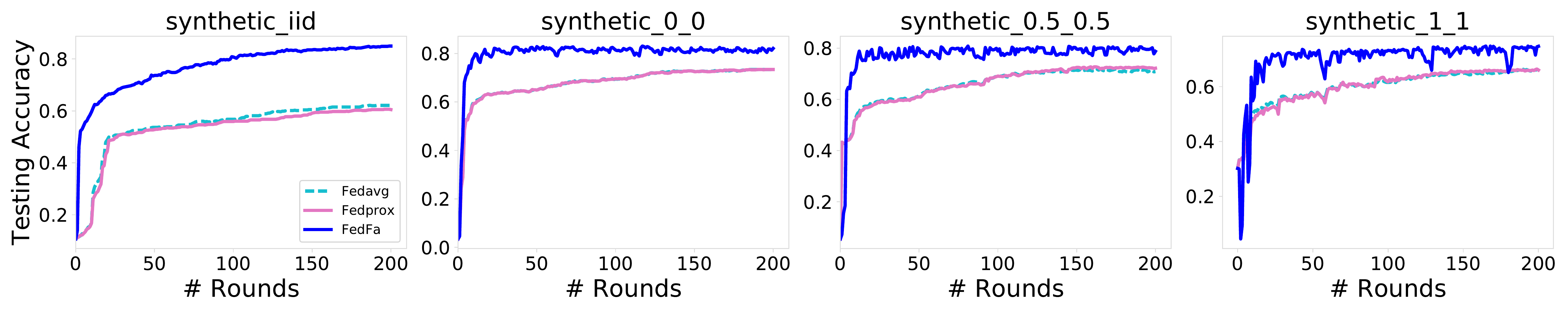}
\label{fig: accuracy-0001-0001}
}

\quad
\subfigure[Training Loss]{
\includegraphics[width=13.7cm]{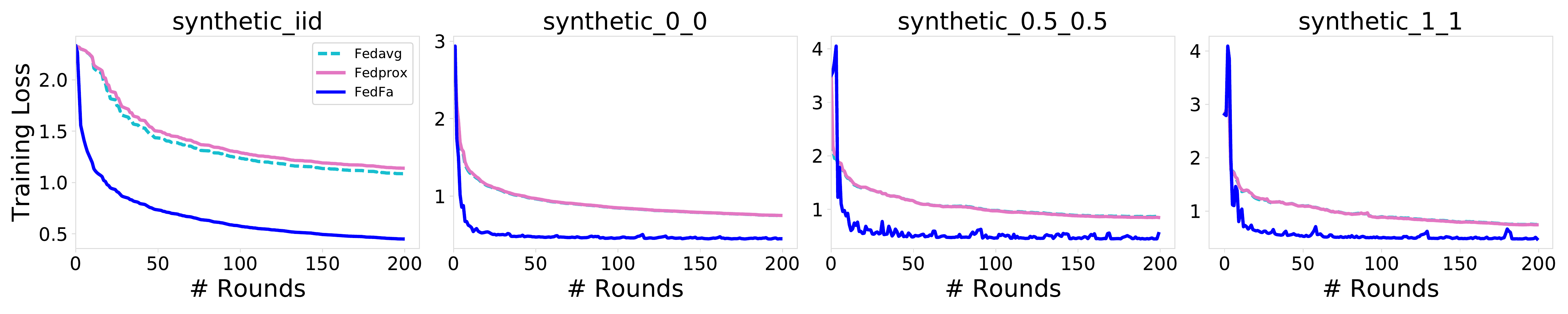}
\label{fig: loss-0001-0001}
}
\caption{The Training Loss and Test Accuracy of Experiments on Synthetic Datasets with Four Different Data Distributions (From left to right, the data are increasingly heterogeneous). The learning rate for FedAvg and FedProx is 0.0001. All settings of FedFa are kept fixed.}
\label{fig: accuracyloss-0001-0001}
\end{figure}

As shown in Figure \ref{fig: accuracyloss-0001-0001}, it is a comparison of test accuracy and loss of FedFa with FedAvg and FedProx at the same learning rate. Obviously, our method is at the advantage of faster convergence. We demonstrate the effectiveness of our proposed double momentum gradient in federated learning through these experiments. The core of the mechanism is to approximate the server-side gradient using the difference between the two rounds of the model on the server-side, which allows the application of the momentum gradient method to the aggregate gradient on the server-side. The client employs the traditional momentum gradient descent method as a means of achieving a double momentum gradient.

Our experiments also show that the momentum gradient method on the server side performs better in heterogeneity problems. The reason is that although the client-side momentum gradient method speeds up its own convergence, it is not friendly to global convergence. Therefore, when the heterogeneity of the data distribution between clients increases, one can consider appropriately decreasing the momentum factor for clients or increasing the round parameter $b$ for the aggregation on the server side.

\subsection{Effects of appropriate weighting strategies}

In our following experiments, we verify that the proposed weighting strategy can lead to a fairer solution for federated datasets. We employ the weighting strategy on the the four synthetic datasets, Femnist and Shakespeare. The accuracy weight $\alpha$ and frequency weight $\beta$ are set to (0.5, 0.5), (0, 1) or (1, 0), respectively.

As shown in the following Table \ref{fair}, our method is compared with FedAvg and Fedprox. The average accuracy, the accuracy of the worst 20\% devices, the accuracy of the best 20\% devices, and the variance of the final accuracy distribution are compared, respectively. In the IID synthetic dataset, our method is slightly inferior to FedAvg, but performs better than the FedProx method. However, in federated networks, the general setup is in the form of Non-IID data distribution. It can be seen that the average accuracy is improved for all datasets except for the IID synthesis dataset. Moreover, the accuracy of the worst and best 20\%  devices is also increased. In addition, the variance of the final accuracy distribution is decreased.

\begin{table*}[!ht]
\centering
\caption{Statistics of the test accuracy distribution for FedAvg, FedProx and FedFa.}\label{fair}
\setlength{\tabcolsep}{6mm}{
\begin{tabular}{c|c|cccl}
\hline
Dataset                              & Method  & Average & Worst 20\% & Best 20\% & Variance \\ \hline
\multirow{3}{*}{Synthetic\_iid}      & FedAvg  & 89.11\% & 78.17\%    & 100.00\%  & 68.04    \\
                                     & FedProx & 71.92\% & 55.46\%    & 86.09\%   & 129.09    \\
                                     & FedFa   & 85.70\% & 71.46\%    & 100.00\%  & 98.74      \\ \hline
\multirow{3}{*}{Synthetic\_0\_0}     & FedAvg  & 54.67\% & 6.81\%     & 98.18\%   & 1112.46   \\
                                     & FedProx & 75.02\% & 37.21\%    & 100.00\%  & 611.10   \\
                                     & FedFa   & \textbf{78.25\%} & \textbf{43.41\%}    & \textbf{100.00\%}  & \textbf{530.27}    \\ \hline
\multirow{3}{*}{Synthetic\_0.5\_0.5} & FedAvg  & 40.24\% & 0.00\%     & 98.18\%   & 1448.46   \\
                                     & FedProx & 67.84\% & 33.60\%    & 100.00\%  & 618.00  \\
                                     & FedFa   & \textbf{73.30\%} & \textbf{41.27\%}    & \textbf{100.00\%}  & \textbf{464.81} \\ \hline
\multirow{3}{*}{Synthetic\_1\_1}     & FedAvg  &54.78\%  &1.38\% &96.85\%  &1069.37  \\
                                     & FedProx &64.75\%  &9.72\% &100.00\% &1088.87   \\
                                     & FedFa   &\textbf{76.88\%}  &\textbf{37.03}\%&\textbf{100.00\%} &\textbf{603.69 }   \\ \hline
\multirow{3}{*}{Femnist}             & FedAvg  &70.96\%  &34.77\%&100.00\% &567.75     \\
                                     & FedProx &72.20\%  &40.50\%&100.00\% &449.83     \\
                                     & FedFa   &\textbf{77.96}\%  &\textbf{48.99}\%&\textbf{100.00}\% &\textbf{368.93}     \\ \hline
\multirow{3}{*}{Sent140}             & FedAvg  &68.51\%  &40.58\% &93.22\%  &356.48       \\
                                     & FedProx &64.71\%  &34.02\% &92.65\%  &431.89      \\
                                     & FedFa   &\textbf{68.83\%}  &\textbf{42.66\%} &\textbf{95.71}\%  &\textbf{319.04}   \\ \hline
\end{tabular}}
\end{table*}


For instance, on the Synthetic\_1\_1 dataset, the heterogeneity between devices is relatively large. The model trained with FedAvg has an accuracy of 1.37\% on the worst 20\% devices. Similarly, on the Synthetic\_1\_1 dataset, the accuracy of the model trained with FedProx on the worst 20\% devices is 9.72\%. This can indicate that as the heterogeneity of data distribution among devices increases, the federated learning training process appears skewed or ignored for some clients, i.e., fairness issues. The model trained with FedFa, on the other hand, has an accuracy of 37.03\% on the worst 20\% devices. FedAvg and FedProx significantly have lower accuracy on the worst 20\% devices than FedFa, but all three algorithms perform well in terms of accuracy on the best 20\% devices.  The results demonstrate that our approach allows clients in federated learning to participate more fairly in the training process, as well as to cope with the heterogeneity problem. The values of $\alpha$ and $\beta$ determine the distribution of training accuracy and the number of clients involved in the training process to be taken more into account. It avoids a federated learning setup with preferences that result in high precision for some clients while others do not participate in training or have low precision.

\begin{figure}[htbp]
\centering
\subfigure{
\includegraphics[width=4.5cm]{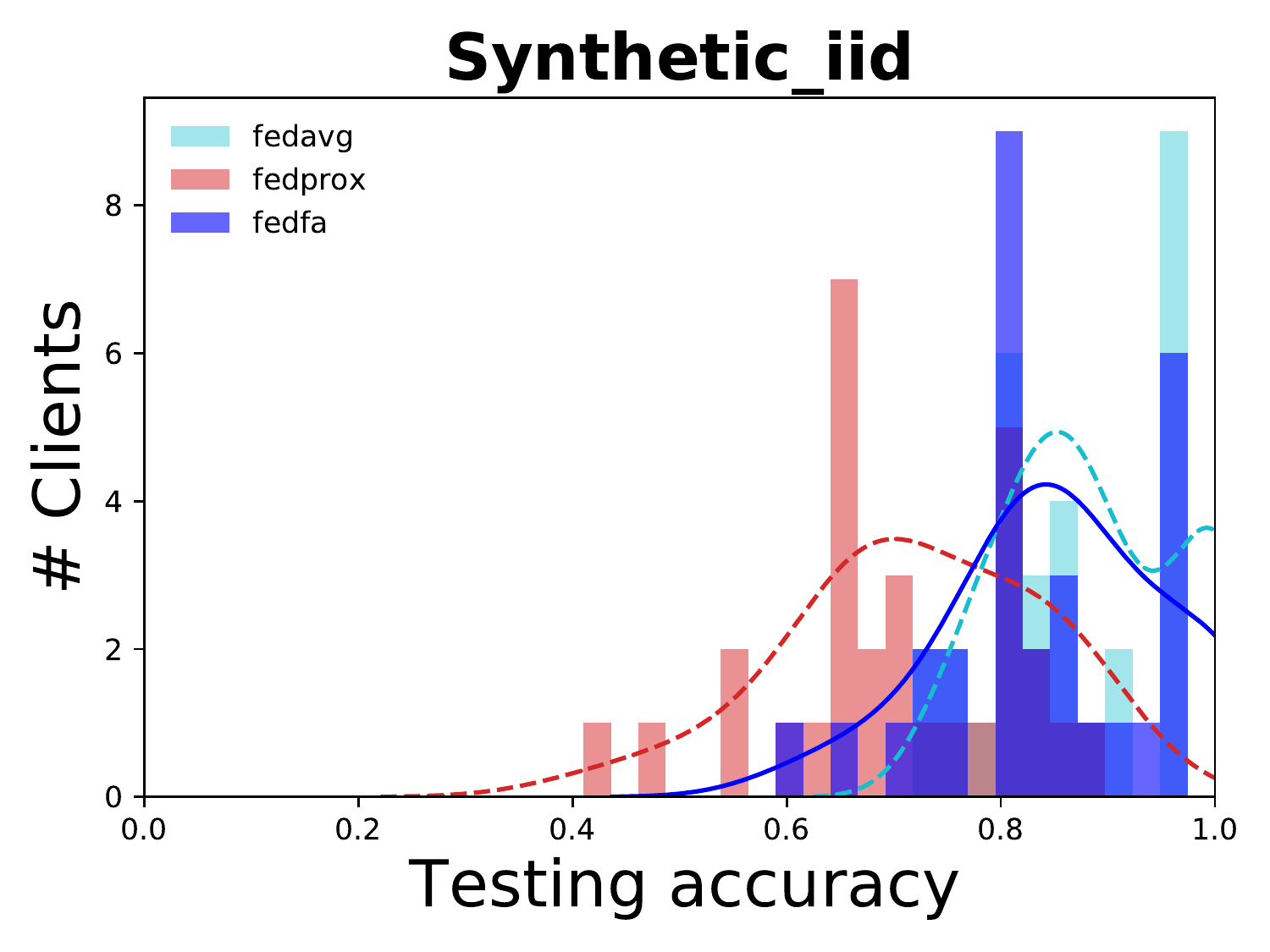}
\label{fig: Synthetic_iid}
}
\subfigure{
\includegraphics[width=4.5cm]{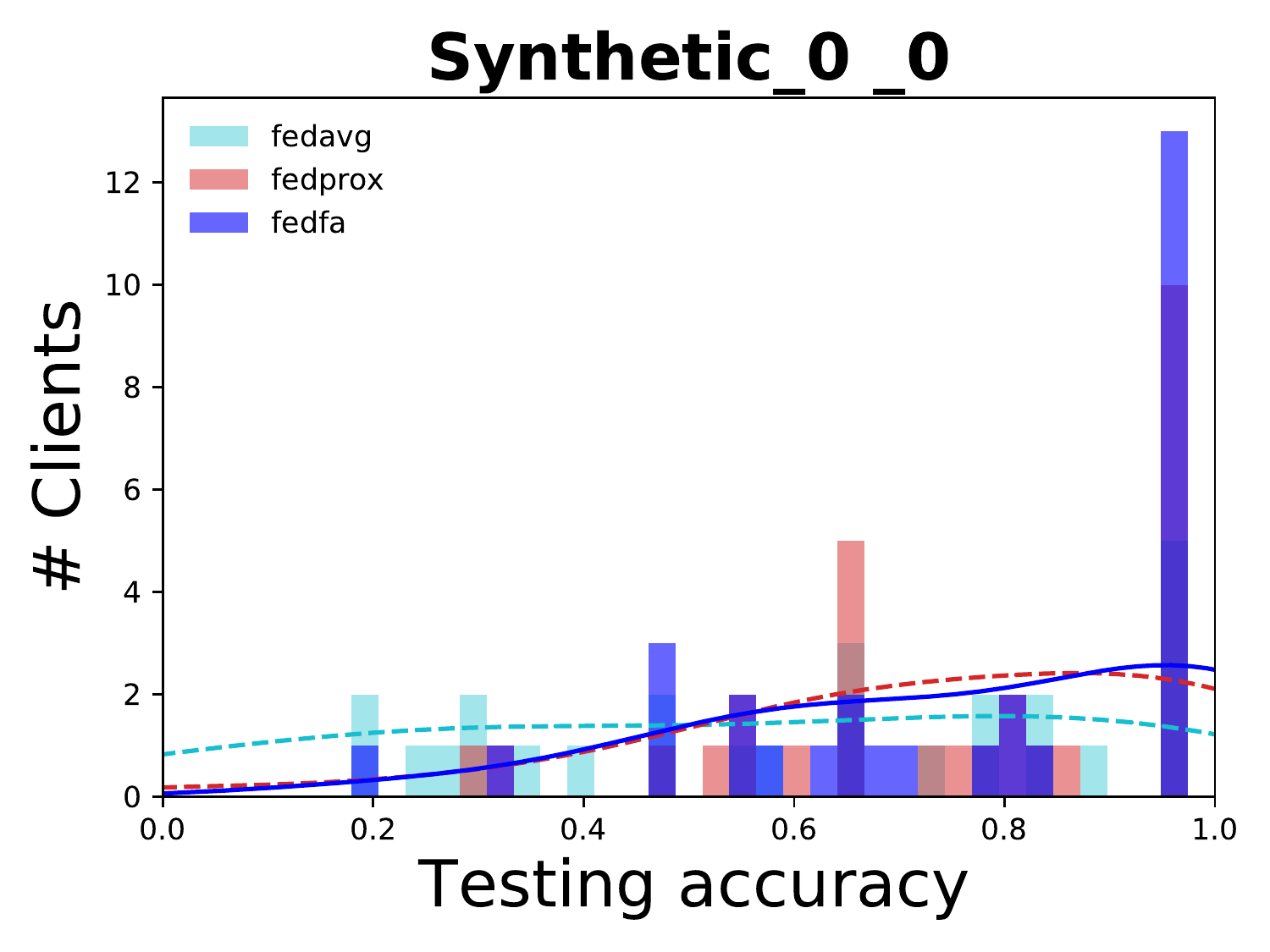}
\label{fig: Synthetic_0_0}
}
\subfigure{
\includegraphics[width=4.5cm]{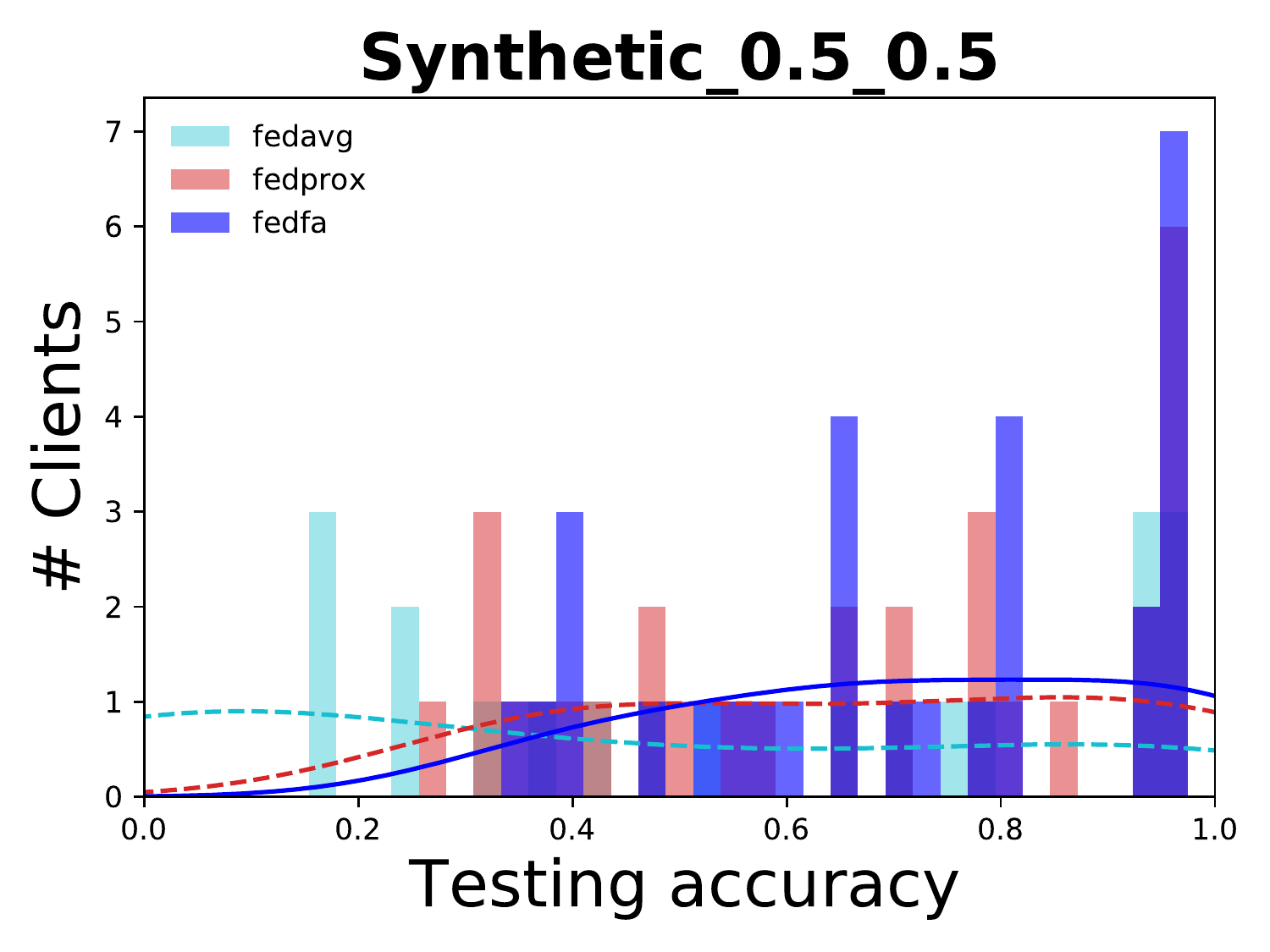}
\label{fig: Synthetic_0.5_0.5}
}
\subfigure{
\includegraphics[width=4.5cm]{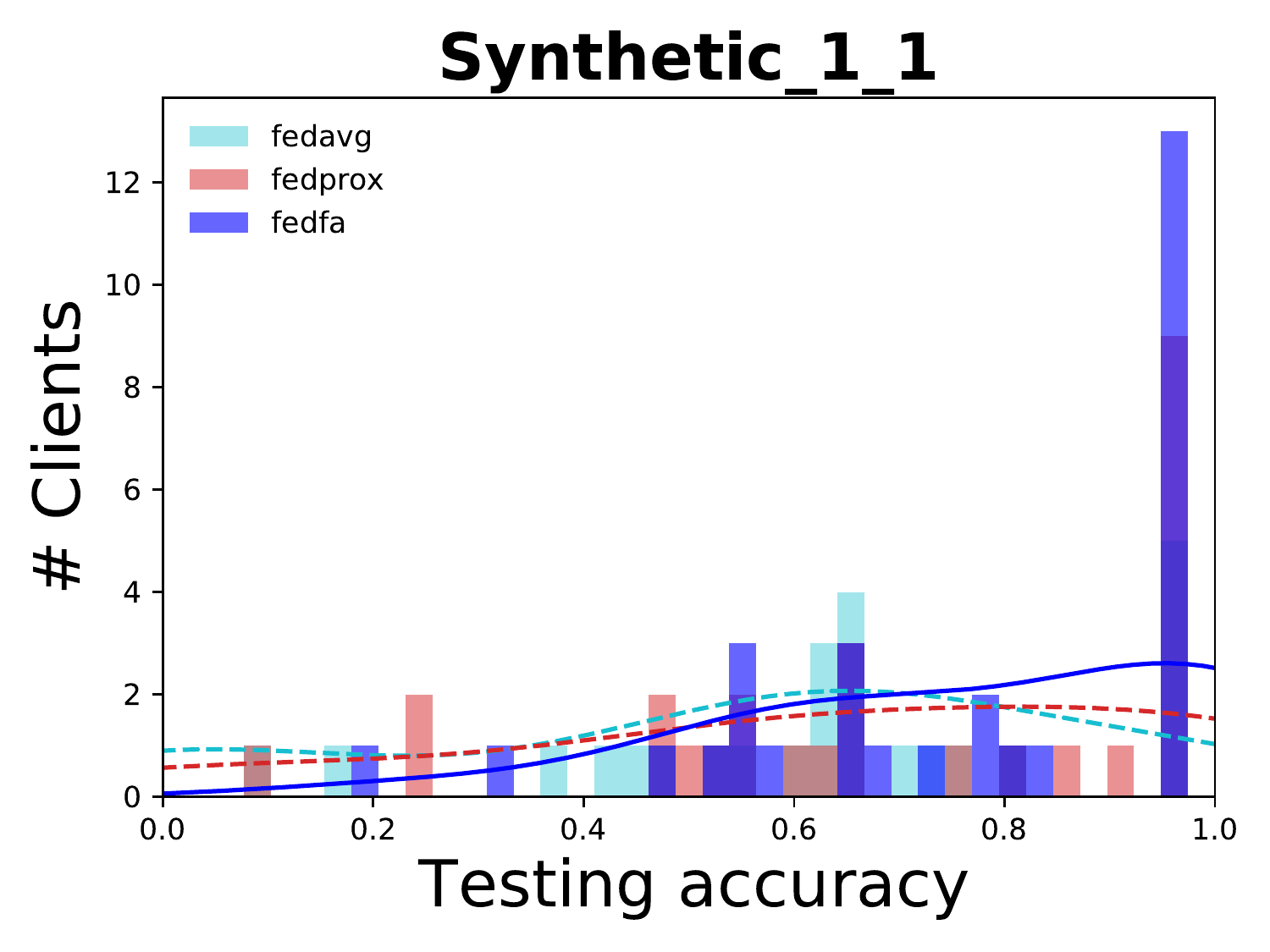}
\label{fig: Synthetic_1_1}
}
\subfigure{
\includegraphics[width=4.5cm]{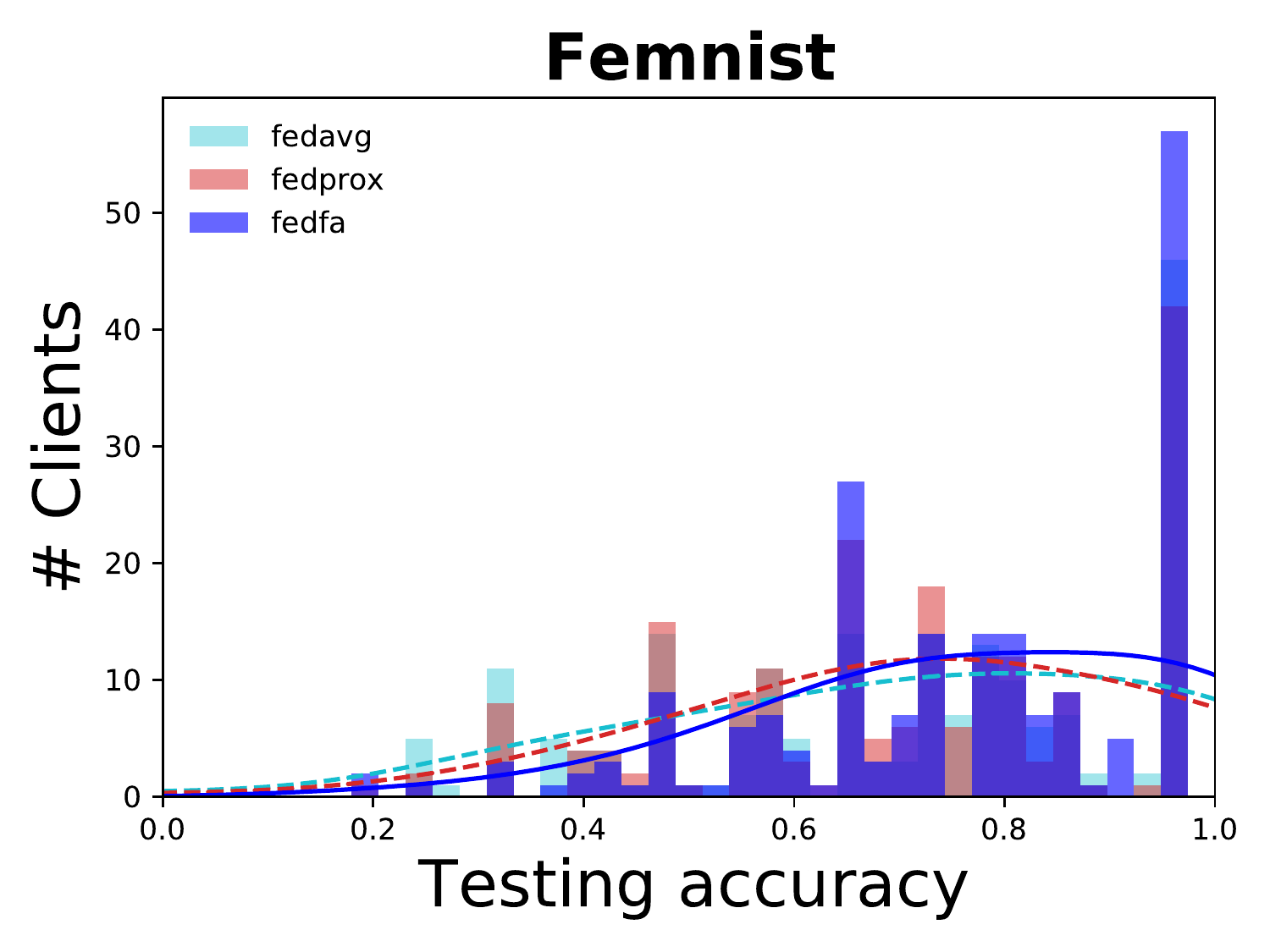}
\label{fig: Femnist}
}
\subfigure{
\includegraphics[width=4.5cm]{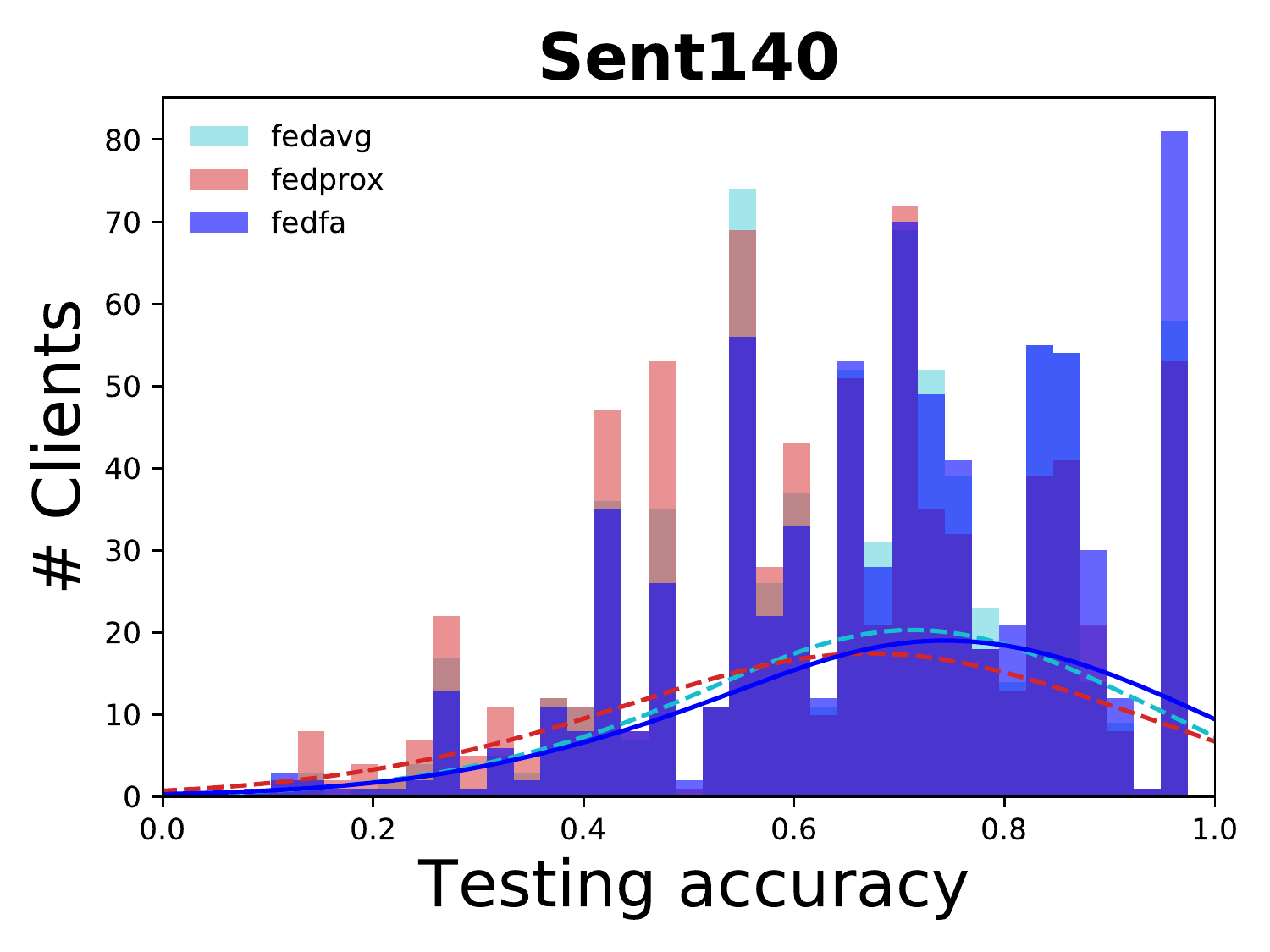}
\label{fig: Sent140}
}
\caption{The distribution of test accuracy for all clients of the four synthetic datasets, Femnsit, and Shakespeare.}
\label{fig: Test accuracy distributions}
\end{figure}

In Figure \ref{fig: Test accuracy distributions} , we present the distribution of clients in terms of final precision for our method and the comparison method. We observe a significant increase in the number of clients with high precision distribution and a significant decrease in the number of clients with low precision or zero precision. Since our method is more fair, it allows more clients to participate in the training process and improve their accuracy.

The basic setup for federated learning often leads to a preference for certain clients in the training process. That is, some clients participate less frequently in the training process, or the accuracy of some clients cannot be improved. Therefore, our approach takes into account the effects of client participation frequency and accuracy to achieve a more fair distribution of accuracy among clients.

\subsection{Discussion of weighted parameters}
In this section, we discuss the turn parameter round $b$ of the momentum gradient method and the parameters $\alpha$ (Accuracy weight) and $\beta$ (Frequency weihgt) of the weighting strategy.

\begin{figure}[htbp]
\centering
\subfigure[Final testing accuracy]{
\includegraphics[width=5.2cm]{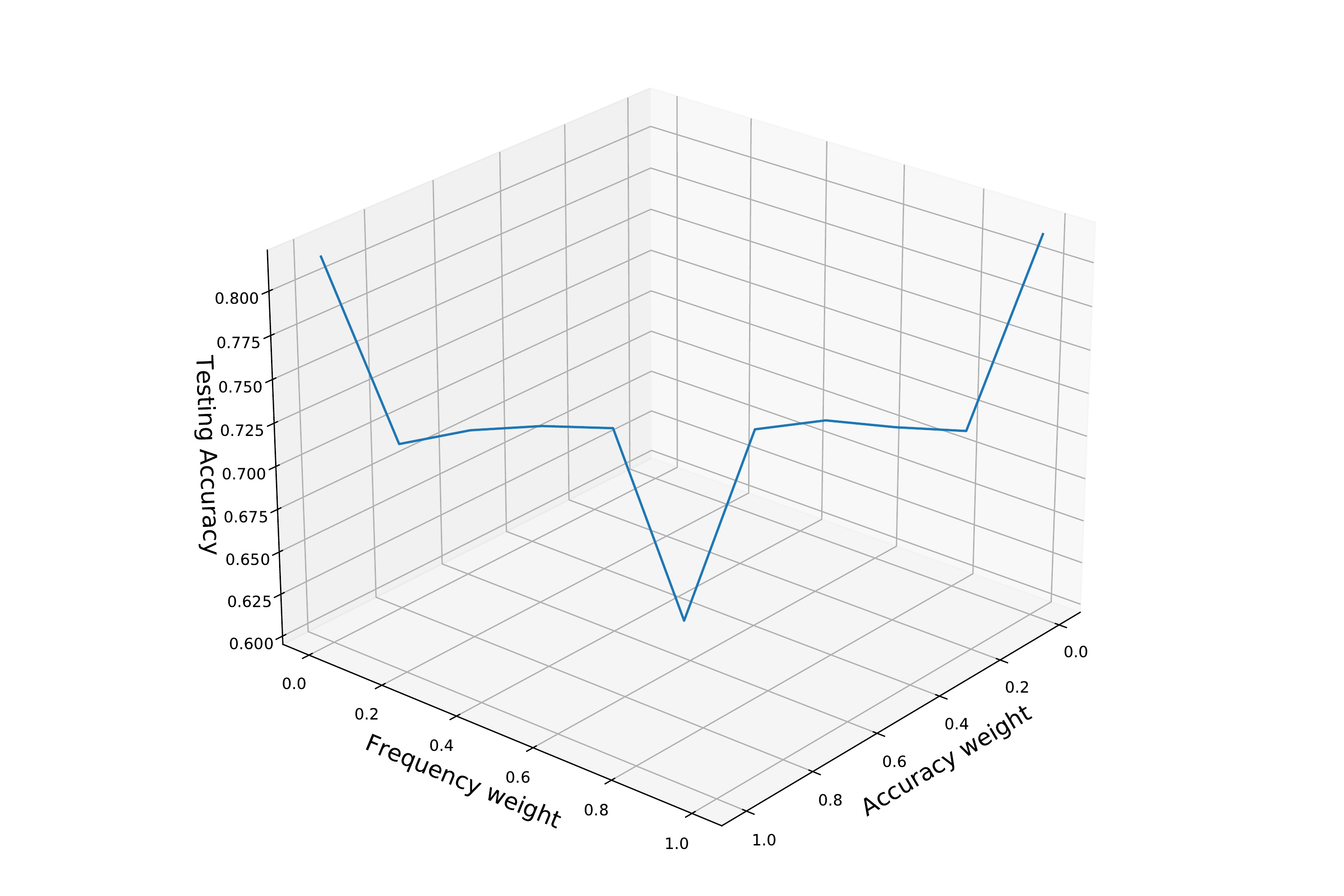}
\label{fig: Parameter_Synthetic_1_1}
}
\subfigure[Variance]{
\includegraphics[width=5.2cm]{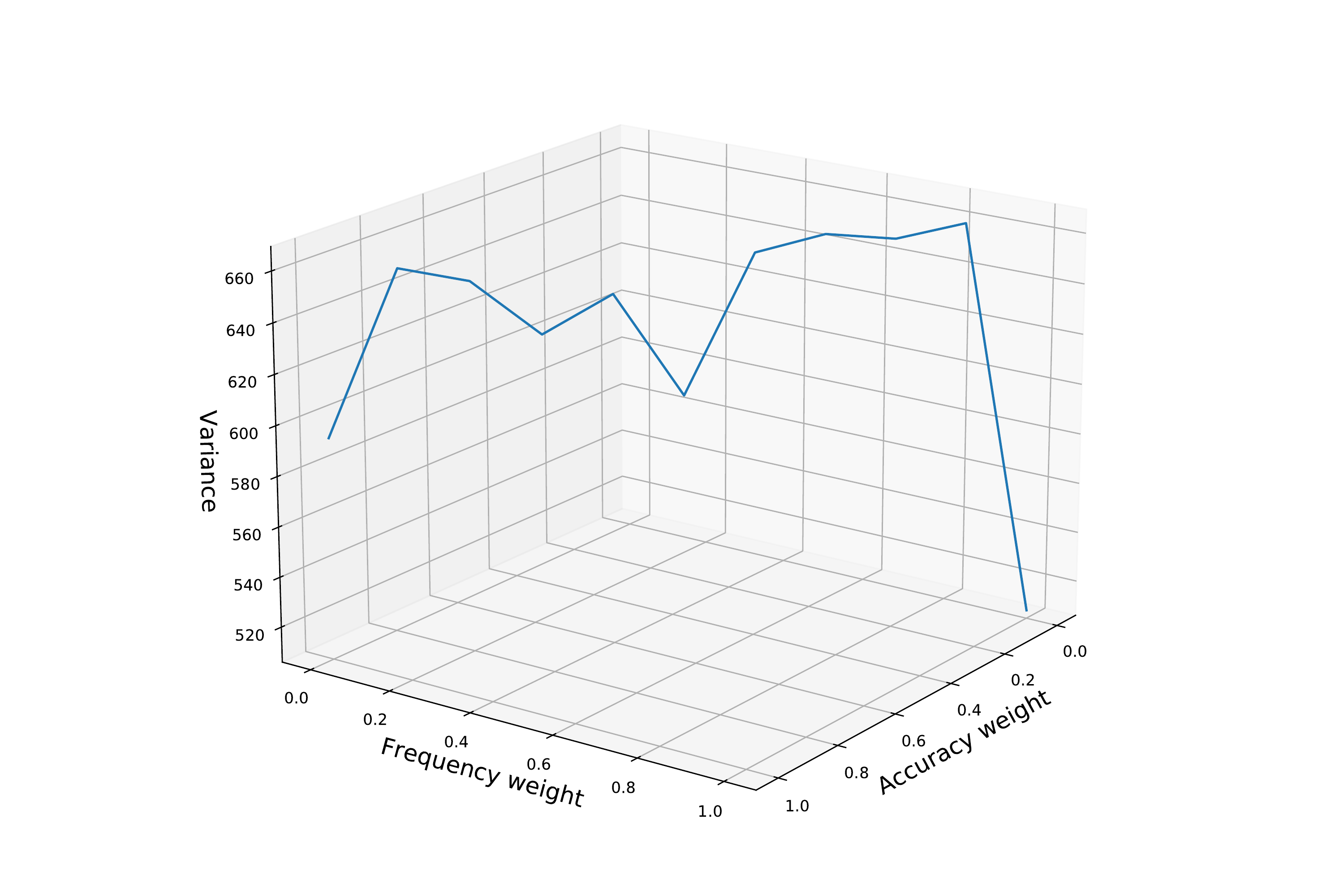}
\label{fig: Variance_Synthetic_1_1}
}
\subfigure[Average accuracy]{
\includegraphics[width=5.2cm]{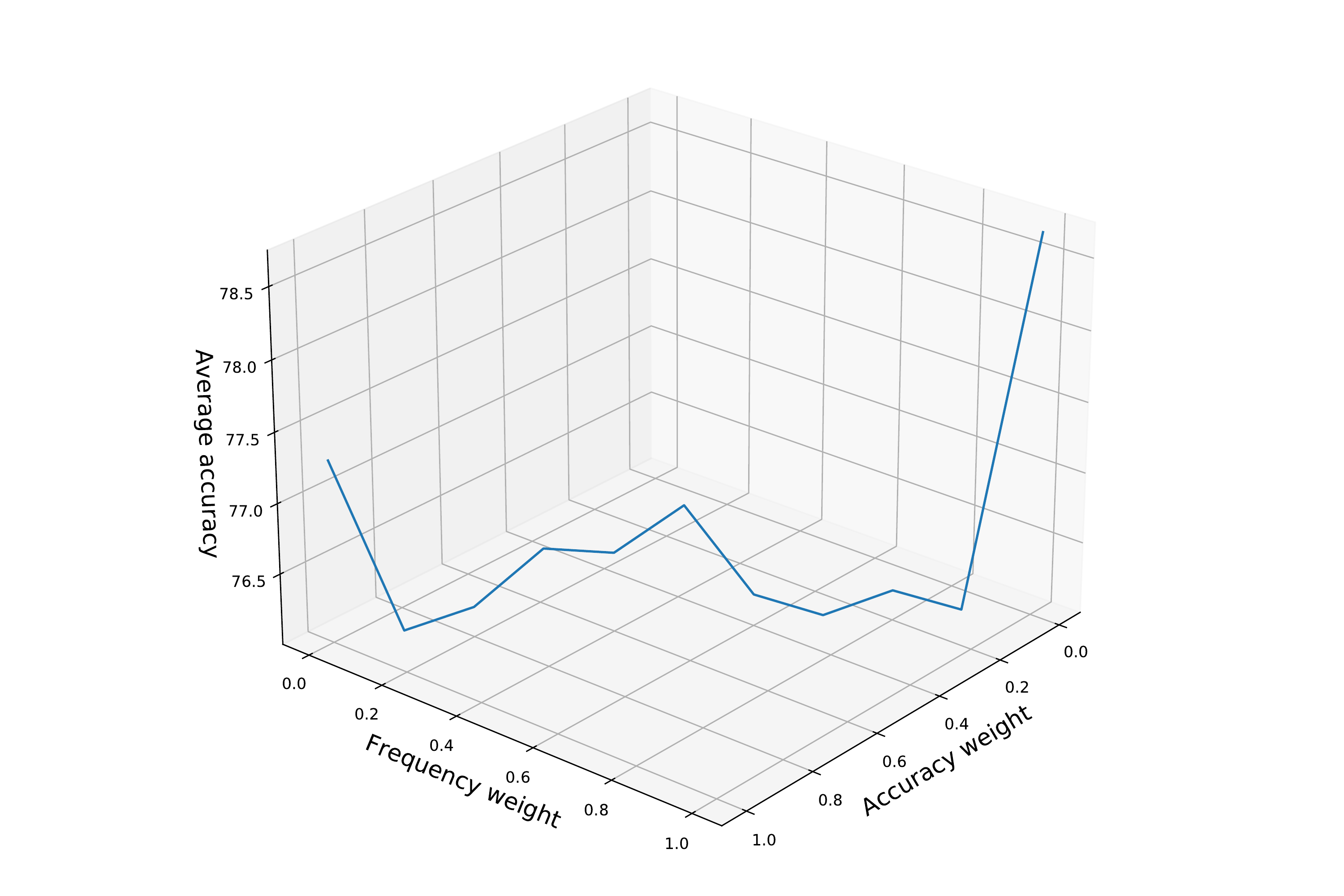}
\label{fig: Average_Synthetic_1_1}
}

\caption{Analysis of the parameters on Synthetic\_1\_1. The frequency weight is the value of $\alpha$ and the accuracy weight is the value of $\beta$. These two parameters determine the final test accuracy, variance and average accuracy.}
\label{fig: Parameter1}
\end{figure}

In order to analyze the effect of the training accuracy weight $\alpha$ and the number of participants weight $\beta$ on the experiment, we increase or decrease these two parameters in steps of 0.1 on the synthetic dataset (Synthetic\_1\_1). However, we have to ensure that $\alpha + \beta =1$. Under this constraint, Figure \ref{fig: Parameter1} shows the effect of parameter values on the experimental results. We can observe the curves of the final accuracy, variance and average accuracy varying with the two parameters. On this dataset, the effect of parameters $\alpha$ and $\beta$ is relatively symmetrical, i.e., $\alpha$ and $\beta$ have a relatively close influence of the final test accuracy.

\begin{figure}[htbp]
\centering
\subfigure[Final testing accuracy]{
\includegraphics[width=5.2cm]{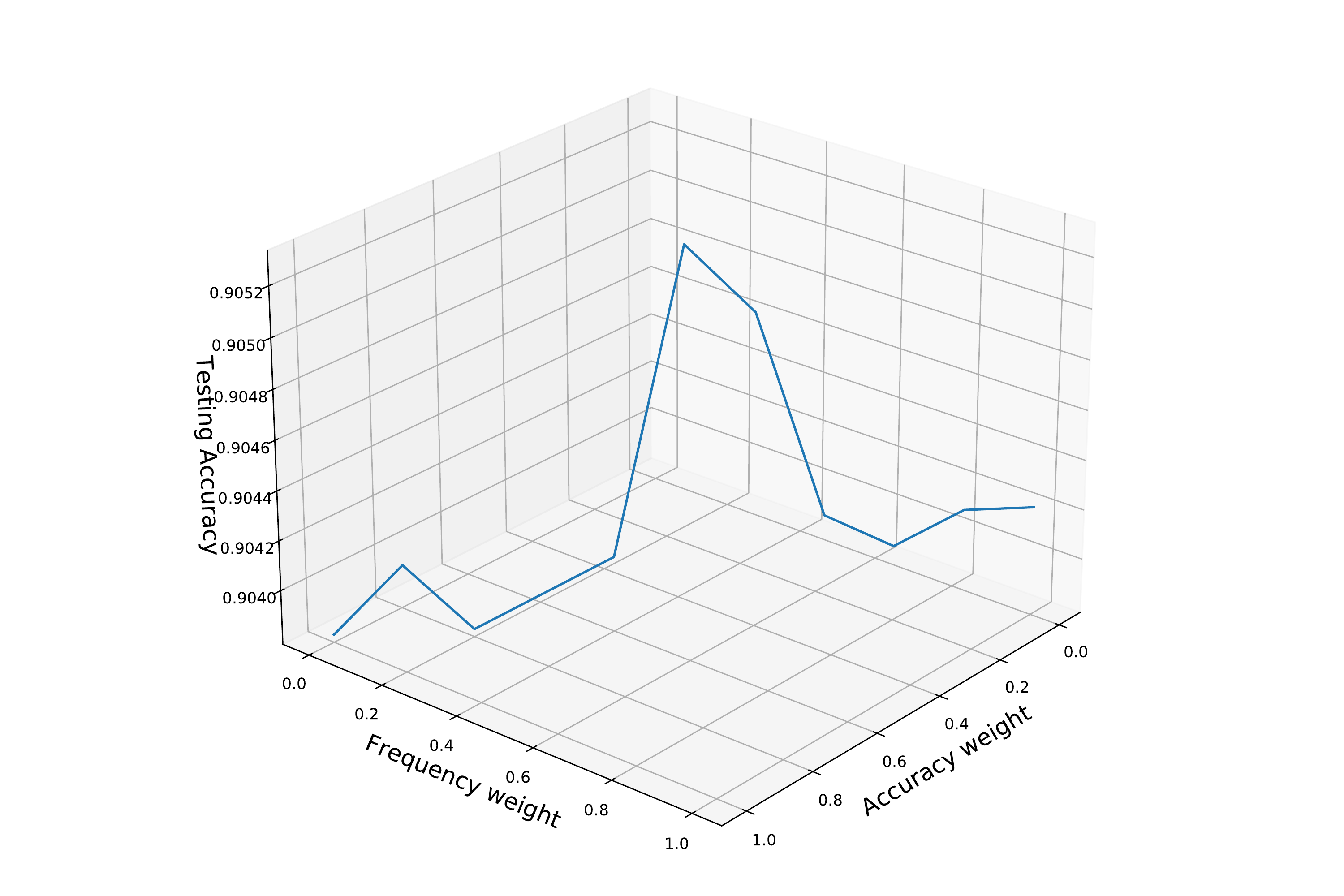}
\label{fig: Parameter_Mnist}
}
\subfigure[Variance]{
\includegraphics[width=5.2cm]{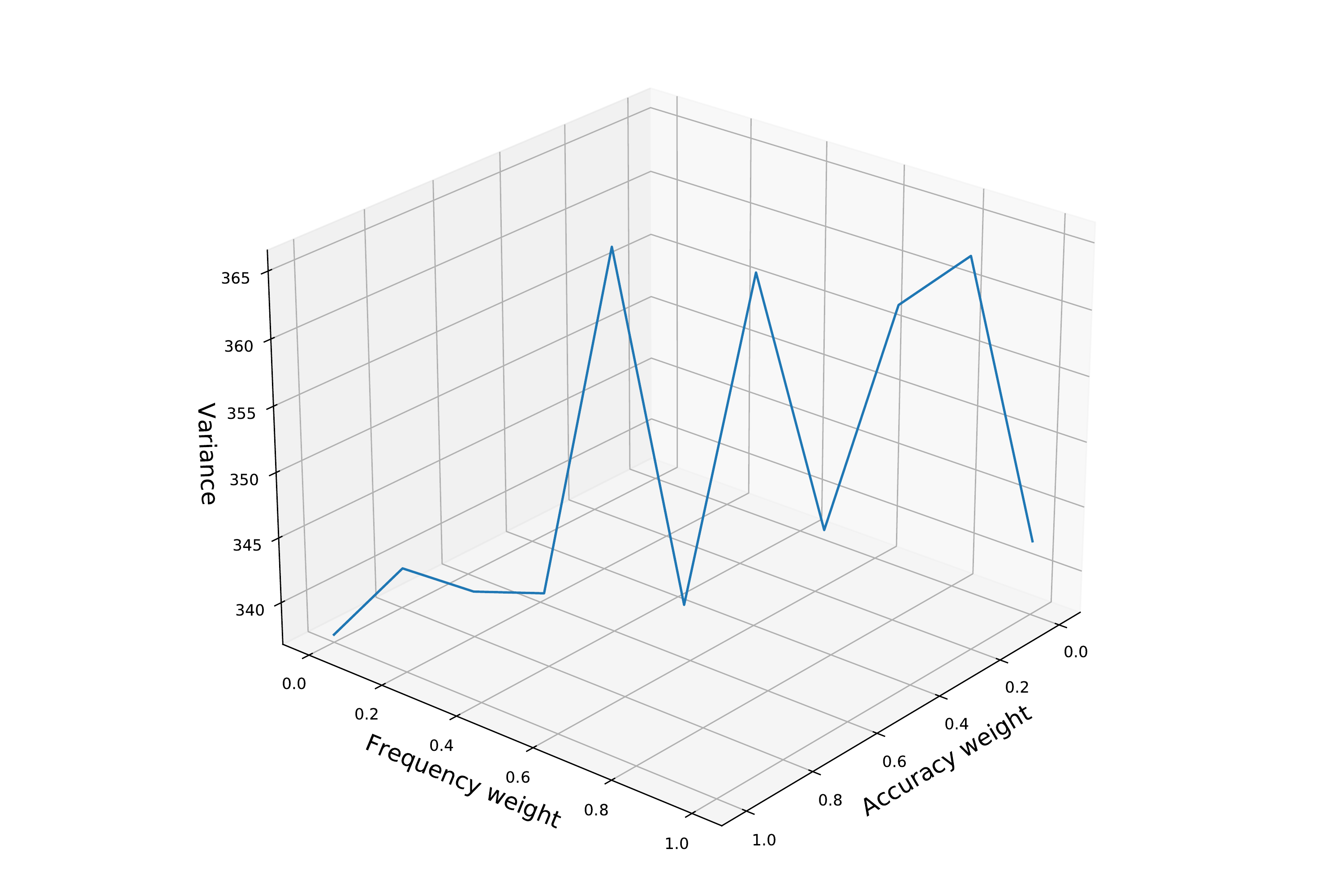}
\label{fig: Variance_Mnist}
}
\subfigure[Average accuracy]{
\includegraphics[width=5.2cm]{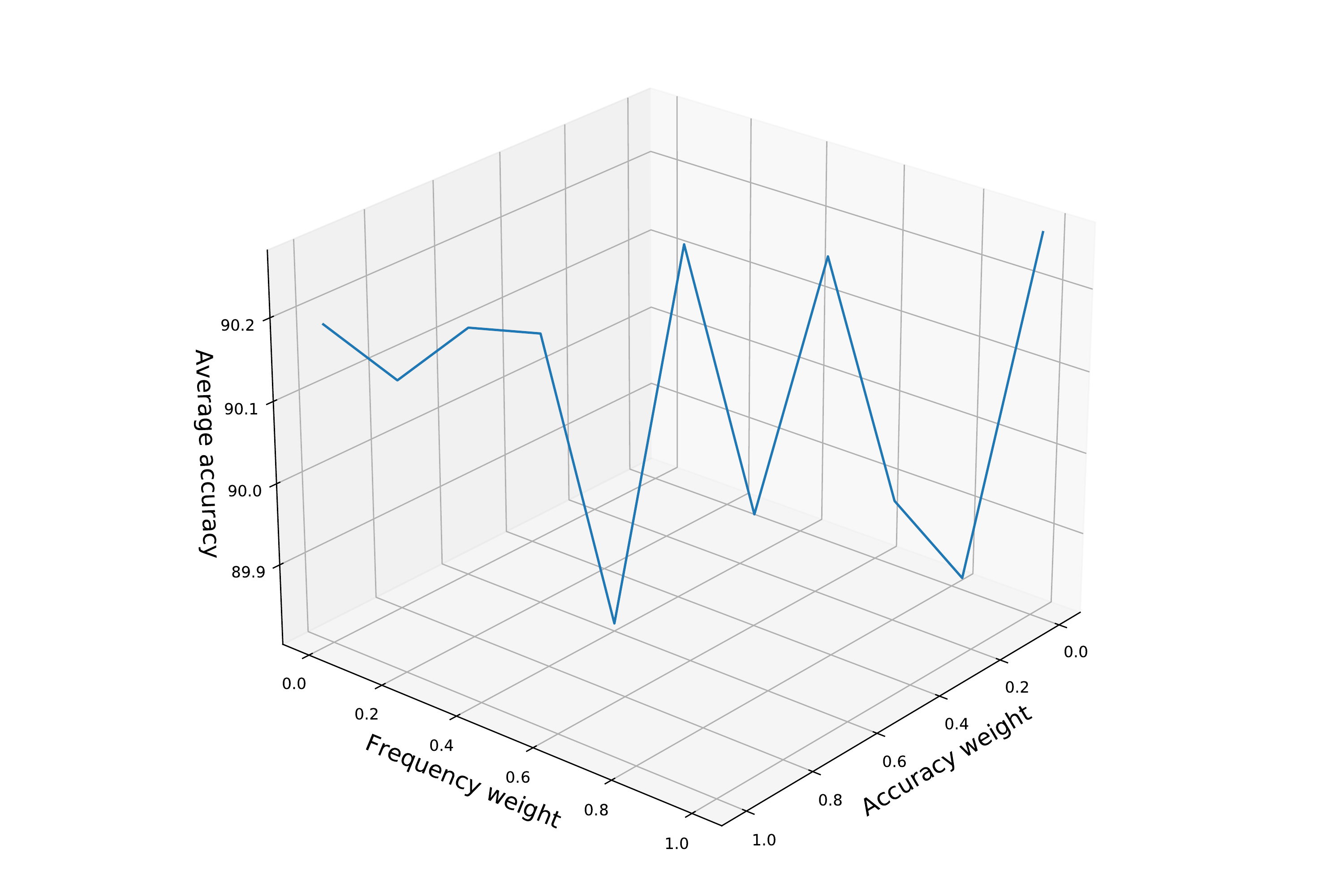}
\label{fig: Average_Mnist}
}

\caption{Analysis of the parameters on Mnist. The frequency weight is the value of $\alpha$ and the accuracy weight is the value of $\beta$. These two parameters determine the final test accuracy, variance and average accuracy.}
\label{fig: Parameter2}
\end{figure}

Figure \ref{fig: Parameter2} illustrates the analysis of the effect of parameters $\alpha$ and $\beta$ on the Mnist dataset. Similarly, we calculate the final accuracy, variance, and average accuracy of the dataset using a step size of 0.1. The figure demonstrates the curves of these metrics varying with the parameters. It is seen that a bias towards one parameter can lead to better experimental results.

The experimental results also demonstrate that the variance between accuracy of the clients is not always small when the average accuracy is large. Therefore, in federated learning, we should focus not only on the average accuracy but also on the variance between accuracy of the clients. In this way, we can make the federated learning framework more fair.

\begin{figure}[htbp]
\centering
\subfigure[Testing Accuracy]{
\includegraphics[width=11cm]{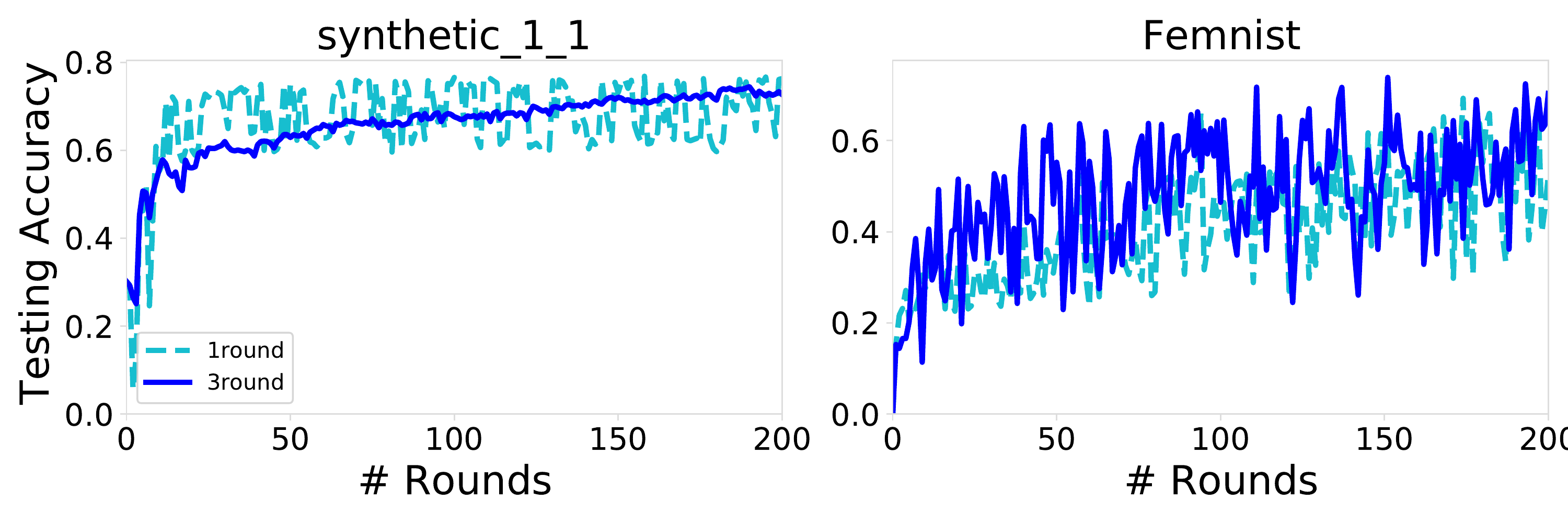}
\label{fig: Testing Accuracy}
}

\subfigure[Training Loss]{
\includegraphics[width=11cm]{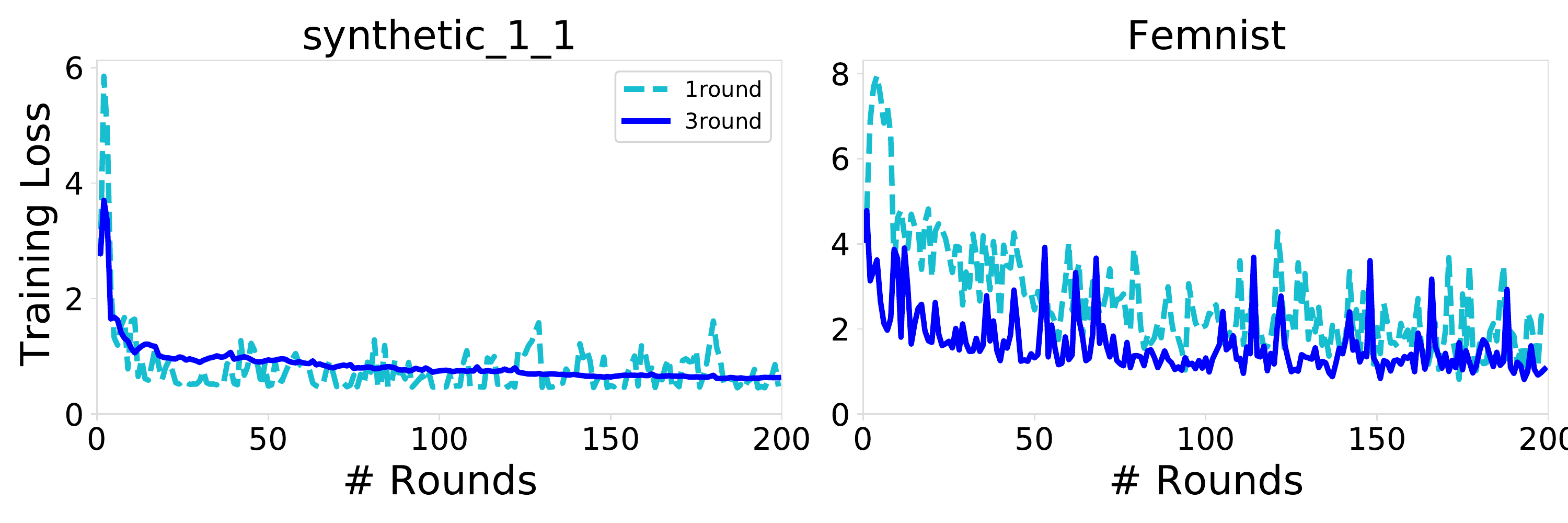}
\label{fig: Training Loss}
}

\caption{Analyze the effects of rounds in Synthetic\_1\_1 and Femnist.}
\label{fig: round}
\end{figure}

In Figure \ref{fig: round}, we conduct experiments using the double momentum gradient method of FedFa on synthetic\_1\_1 and Femnist. We compare the results of our experiments using the momentum gradient method after 1 round and 3 rounds of aggregation on the server side. We have verified that the momentum gradient method can be used at each round when the server-side performs the convergence gradient. It can be seen from Figure \ref{fig: round} that when the number of rounds of the server is taken as 3, it will be more stable than using the momentum gradient method for each round.

It is also possible to use the momentum gradient method on the server-side after round $b$. Since the problem of heterogeneity in the data distribution of these clients leads to different gradient optimization directions for each client. Therefore, we can wait for the server to collect the gradients of round $b$ before using the momentum gradient method, which can better consider the influence of historical gradients and reduce the oscillation.

\section{Conclusions}
In this paper, we proposed a novel federated optimization algorithm, FedFa, which combines a double momentum gradient and a weighting strategy. The double momentum gradient takes into account the influence of historical gradient information at both client and server sides to improve the convergence of the algorithm. The weighting strategy aggregates the weights according to the training accuracy and the number of participants to create a fairer and more accurate federated learning algorithm. Through extensive experiments on federated datasets, we validated that our proposed approach significantly improves the convergence and fairness of federated learning in heterogeneous networks compared to existing benchmarks. In future research, we will develop new federated learning algorithms that can better adapt to the challenge of heterogeneity among clients, thus further improving learning performance and reducing communication costs.

\section*{Acknowledgments}
This research was supported by the National Key R$\&$D Program of China (2019YFB2101802) and the National Natural Science Foundation of China (No. 61773324).

\section*{References}
\renewcommand \baselinestretch{0.5} \selectfont

\bibliographystyle{elsarticle-num} 

\bibliography{FedFa}
\end{CJK*}
\end{document}